\documentclass{article}

\usepackage[nonatbib, final]{neurips_2021}

\usepackage{times}
\usepackage{epsfig}
\usepackage{graphicx}
\usepackage{amsmath}
\usepackage{amssymb}

\usepackage[utf8]{inputenc} %
\usepackage[T1]{fontenc}    %
\usepackage{hyperref}       %
\usepackage{url}            %
\usepackage{booktabs}       %
\usepackage{amsfonts}       %
\usepackage{nicefrac}       %
\usepackage{microtype}      %
\usepackage{xcolor}         %

\title{Canonical Capsules: \\ Self-Supervised Capsules in Canonical Pose}

\author{%
    Weiwei Sun$^{1,4,*}$\And
    Andrea Tagliasacchi$^{2,3,*}$\And 
    Boyang Deng$^{3}$\And
    Sara Sabour$^{2,3}$\And
    Soroosh Yazdani$^{3}$ $\qquad$
    Geoffrey Hinton$^{2,3}$ $\qquad$
    Kwang Moo Yi$^{1,4}$\\[.2in]
    $^1$University of British Columbia, \hspace{3pt}
    $^2$University of Toronto, \hspace{3pt} \\
    $^3$Google Research, \hspace{3pt}
    $^4$University of Victoria, \hspace{3pt} $^*$equal contributions
    \\[.2in]
    \url{https://canonical-capsules.github.io}
}

\usepackage{overpic}
\usepackage{microtype}
\usepackage{enumitem}
\usepackage{overpic}
\usepackage{placeins}
\usepackage{booktabs}
\usepackage{bbding}
\usepackage{comment}
\usepackage{subfigure}
\usepackage[normalem]{ulem}
\usepackage{multicol}
\usepackage{wrapfig}

\usepackage[resetlabels,labeled]{multibib}
\newcites{A}{Additional References}

\usepackage{color}
\definecolor{turquoise}{cmyk}{0.65,0,0.1,0.3}
\definecolor{purple}{rgb}{0.65,0,0.65}
\definecolor{dark_green}{rgb}{0, 0.5, 0}
\definecolor{orange}{rgb}{0.8, 0.6, 0.2}
\definecolor{red}{rgb}{0.8, 0.2, 0.2}
\definecolor{darkred}{rgb}{0.6, 0.1, 0.05}
\definecolor{blueish}{rgb}{0.0, 0.3, .6}
\definecolor{light_gray}{rgb}{0.7, 0.7, .7}
\definecolor{pink}{rgb}{1, 0, 1}
\definecolor{greyblue}{rgb}{0.25, 0.25, 1}

\renewcommand{\paragraph}[1]{\vspace{.0em}\noindent\textbf{#1}.}

\usepackage{blindtext}

\newcommand{\Figure}[1]{Figure~\ref{fig:#1}}

\newcommand{\Table}[1]{Table~\ref{tbl:#1}}
\newcommand{\eq}[1]{\eqref{eq:#1}}

\newcommand{\Section}[1]{Section~\ref{sec:#1}}

\DeclareMathOperator*{\argmin}{arg\,min}

\newcommand{\loss}[1]{\mathcal{L}_\text{#1}}

\newcommand{\real}{\mathbb{R}}
\newcommand{\mR}{\mathbf{R}}
\newcommand{\mT}{\mathbf{T}}
\newcommand{\vt}{\mathbf{t}}
\newcommand{\bA}{\mathbf{A}}
\newcommand{\bW}{\mathbf{W}}

\newcommand{\bmu}{\boldsymbol{\mu}}
\newcommand{\bsigma}{\boldsymbol{\sigma}}

\newcommand{\pose}{\boldsymbol{\theta}}
\newcommand{\descriptor}{\boldsymbol{\beta}}
\newcommand{\featuremap}{\mathbf{F}}
\newcommand{\attention}{\mathbf{A}}
\newcommand{\points}{\mathbf{P}}
\newcommand{\acne}{\mathcal{E}}
\newcommand{\decoder}{\mathcal{D}}
\newcommand{\regressor}{\mathcal{K}}

\newcommand{\CIRCLE}[1]{\raisebox{.5pt}{\footnotesize \textcircled{\raisebox{-.6pt}{#1}}}}

\newcommand{\SupplementaryMaterial}{\texttt{supplementary material}\xspace}
\newcommand{\capsnet}{\mbox{3D-PointCapsNet}~\cite{zhao20193d}\xspace}
\newcommand{\atlasnet}{AtlasNetV2~\cite{deprelle2019learning}\xspace}
\newcommand{\compass}{Compass~\cite{spezialetti2020learning}\xspace}
\newcommand{\deepgmr}{DeepGMR~\cite{yuan2020deepgmr}\xspace}

\newcommand{\SE}[1]{$\mathbf{SE}(#1)$}
\newcommand{\SO}[1]{$\mathbf{SO}(#1)$}

\makeatletter
\DeclareRobustCommand\onedot{\futurelet\@let@token\@onedot}
\def\@onedot{\ifx\@let@token.\else.\null\fi\xspace}

\def\eg{\emph{e.g}\onedot} 
\def\ie{\emph{i.e}\onedot}

\makeatother
\usepackage{multirow}
\usepackage{caption}

\begin{document}

\maketitle

\begin{abstract}
We propose a self-supervised capsule architecture for 3D point clouds.
We compute capsule decompositions of objects through permutation-equivariant attention, and self-supervise the process by training with pairs of randomly rotated objects.
Our key idea is to aggregate the attention masks into semantic keypoints, and use these to supervise a decomposition that satisfies the capsule invariance/equivariance properties.
This not only enables the training of a semantically consistent decomposition, but also allows us to learn a canonicalization operation that enables object-centric reasoning.
To train our neural network we require neither classification labels nor manually-aligned training datasets.
Yet, by learning an object-centric representation in a self-supervised manner, our method outperforms the state-of-the-art on 3D point cloud reconstruction, canonicalization, and unsupervised classification.
\end{abstract}

\newcommand{\footnoteautoencoding}{Auto-encoding is also at times referred to as ``reconstruction'' or ``shape-space'' learning.}

\section{Introduction}
\label{sec:intro}
Understanding objects is one of the core problems of computer vision~\cite{sift,dpm,yolo}.
While this task has traditionally relied on large annotated datasets~\cite{simonyan2014very,he2016deep}, unsupervised approaches that utilize self-supervision~\cite{chen2020simple} have emerged to remove the need for labels.
Recently, researchers have attempted to extend these methods to work on 3D point clouds~\cite{pointcontrast}, but the field of unsupervised 3D learning remains relatively uncharted.
Conversely, researchers have been extensively investigating 3D deep representations for shape auto-encoding\footnote{\footnoteautoencoding}~\cite{yang2018foldingnet,groueix2018papier,mescheder2019occupancy,genova2019deep}, making one wonder whether these discoveries can now benefit from unsupervised learning for tasks \textit{other} than auto-encoding.

Importantly, these recent methods for 3D deep representation learning are not entirely unsupervised.
Whether using point clouds~\cite{yang2018foldingnet}, meshes~\cite{groueix2018papier}, or implicits~\cite{mescheder2019occupancy}, they owe much of their success to the bias within the dataset that was used for training.
Specifically, all 3D models in the popular ShapeNet~\cite{shapenet} dataset are ``object-centric'' -- they are pre-canonicalized to a unit bounding box, and, even more importantly, with an orientation that synchronizes object semantics to Euclidean frame axes (e.g. airplane cockpit is always along $+y$, car wheels always touch $z=0$).
Differentiable 3D decoders are heavily affected by the consistent alignment of their output with an Euclidean frame~\cite{deng2019neural,genova2019deep} as local-to-global transformations \textit{cannot} be easily learnt by fully connected layers.
As we will show in \Section{reconstruction}, these methods fail in the absence of pre-alignment, even when data augmentation is used.
A concurrent work~\cite{spezialetti2020learning} also recognizes this problem and proposes a separate learnt canonicalizer, which is shown helpful in downstream classification tasks.

In this work, we leverage the modeling paradigm of capsule networks~\cite{hinton2011transforming}.
In capsule networks, a scene is perceived via its decomposition into \textit{part} hierarchies, and each part is represented with a~(pose, descriptor) pair:
\CIRCLE{1} The capsule \emph{pose} specifies the frame of reference of a part, and hence should be transformation equivariant;
\CIRCLE{2} The capsule \emph{descriptor} specifies the appearance of a part, and hence should be transformation invariant.
Thus, one does not have to worry how the data is oriented or translated, as these changes can be encoded within the capsule representation.

We introduce \textit{Canonical \textbf{Capsules}}, a novel capsule architecture to compute a K-part decomposition of a point cloud.
We train our network by feeding pairs of a randomly rotated/translated copies of the same shape (i.e.~siamese training) hence removing the requirement of pre-aligned training datasets.
We then decompose the point cloud by assigning each point into one of the~K parts via attention, which we aggregate into~K keypoints.
Equivariance is then enforced by requiring the two keypoint sets to only differ by the known (relative) transformation (i.e.~a form of self-supervision).
For invariance, we simply ask that the descriptors of each keypoint of the two instances match.

With \textit{\textbf{Canonical} Capsules}, we exploit our decomposition to recover a canonical frame that allows unsupervised ``object-centric'' learning of 3D deep representations \textit{without} requiring a semantically aligned dataset.
We achieve this task by regressing \textit{canonical} capsule poses from capsule descriptors via a deep network, and computing a canonicalizing transformation by solving a shape-matching problem~\cite{shapematching}.
This not only allows more effective shape auto-encoding, but our experiments confirm this results in a latent representation that is more effective in unsupervised classification tasks.
Note that, like our decomposition, our canonicalizing transformations are also learnt in a self-supervised fashion, by only training on randomly transformed point clouds.

\paragraph{Contributions}
In summary, in this paper we:
\vspace{-0.5em}
\begin{itemize}[leftmargin=*]
    \setlength\itemsep{-.3em}
    \item propose an architecture for 3D self-supervised learning based on capsule decomposition;
    \item enable object-centric unsupervised learning by introducing a learned canonical frame of reference;
    \item achieve state-of-the-art performance 3D point cloud auto-encoding/reconstruction, canonicalization, and unsupervised classification.
\end{itemize}

\section{Related works}

Convolutional Neural Networks lack equivariance to rigid transformations, despite their pivotal role in describing the structure of the 3D scene behind a 2D image.
One promising approach to overcome this shortcoming is to add equivariance under a group action in each layer~\cite{thomas2018tensor,deng2021vn}. 
In our work, we remove the need for a global \SE{3}-equivariant network by canonicalizing the input.

\paragraph{Capsule Networks}
Capsule Networks~\cite{hinton2011transforming} have been proposed to overcome this issue towards a relational and hierarchical understanding of natural images.
Of particular relevance to our work, are methods that apply capsule networks to 3D input data~\cite{zhao20193d,zhao2019quaternion,srivastava2019geometric}, but note these methods are not unsupervised, as they either rely on classification supervision~\cite{zhao2019quaternion}, or on datasets that present a significant inductive bias in the form of pre-alignment~\cite{zhao20193d}.
In this paper, we take inspiration from the recent Stacked Capsule Auto-Encoders~\cite{kosiorek2019stacked}, which shows how capsule-style reasoning can be effective as long as the \textit{primary} capsules can be trained in a self-supervised fashion (i.e.~via reconstruction losses).
The natural question, which we answer in this paper, is \textit{``How can we engineer networks that generate 3D primary capsules in an unsupervised fashion?''}

\paragraph{Deep 3D representations}
Reconstructing 3D objects requires effective inductive biases about 3D vision \textit{and} 3D geometry.
When the input is images, the core challenge is how to encode \textit{3D projective geometry} concepts into the model.
This can be achieved by explicitly modeling multi-view geometry~\cite{kar2017learning}, by attempting to learn it~\cite{fan2017point}, or by hybrid solutions~\cite{yan2016perspective}.
But even when input is 3D, there are still significant challenges.
It is still not clear which is the 3D \textit{representation} that is most amenable to deep learning.
Researchers proposed the use of meshes~\cite{wang2018pixel2mesh, litany2018deformable}, voxels~\cite{wang2017cnn, wang2018adaptive}, surface patches~\cite{groueix2018papier, deprelle2019learning, deng2020better}, and implicit functions~\cite{park2019deepsdf, mescheder2019occupancy, chen2019learning}.
Unfortunately, the importance of geometric structures (i.e.~\textit{part-to-whole} relationships) is often overlooked.
Recent works have tried to close this gap by using part decomposition consisting of oriented boxes~\cite{tulsiani2017learning}, ellipsoids~\cite{genova2019learning, genova2019deep}, convex polytopes~\cite{deng2020cvxnet}, and grids~\cite{chabra2020deep}.
However, as previously discussed, most of these still heavily rely on a pre-aligned training dataset; our paper attempts to bridge this gap, allowing learning of \textit{structured} 3D representations \textit{without} requiring pre-aligned data.

\paragraph{Canonicalization}
One way to circumvent the requirement of pre-aligned datasets is to rely on methods capable of registering a point cloud into a canonical frame.
The recently proposed CaSPR~\cite{rempe2020caspr} fulfills this premise,
but requires ground-truth canonical point clouds in the form of normalized object coordinate spaces~\cite{wang2019normalized} for supervision.
Similarly,~\cite{gu2020weakly} regresses each view's pose relative to the canonical pose, but still requires weak annotations in the form of multiple partial views.
C3DPO~\cite{novotny2019c3dpo} learns a canonical 3D frame based on 2D input keypoints.
In contrast to these methods, our solution is completely self-supervised.
The concurrent Compass~\cite{spezialetti2020learning} also learns to canonicalize in a self-supervised fashion, but as the process is not end-to-end, this results in a worse performance than ours, as it will be shown in~\Section{canonicalization}.

\paragraph{Registration}
Besides canonicalization, many \textit{pairwise} registration techniques based on deep learning have been proposed~(e.g.~\cite{wang2019deep,yuan2020deepgmr}), even using semantic keypoints and symmetry to perform the task~\cite{fernandez2020unsupervised}.
These methods typically register a \textit{pair} of instances from the same class, but lack the ability to \textit{jointly} and consistently register all instances to a shared canonical frame.

\begin{figure*}
    \begin{center}
        \includegraphics[width=\linewidth]{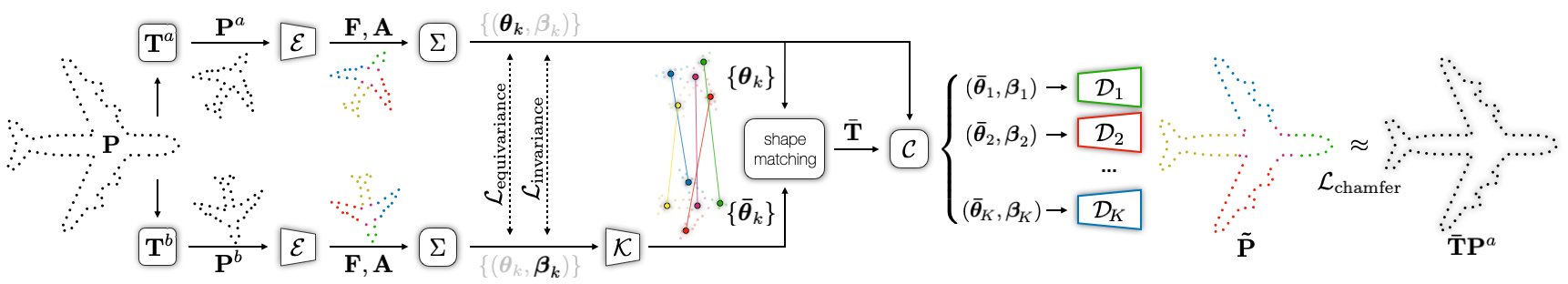}
    \end{center}
    \caption{
        \textbf{Framework -- }
        We learn a capsule encoder for 3D point clouds by relating the decomposition result of two random rigid transformations $\mT^a$ and $\mT^b$, of a given point cloud, \ie, a Siamese training setup.
        We learn the parameters of an encoder $\acne$, a per-capsule decoder $\decoder_k$, as well as a network that represents a learnt canonical frame $\regressor$.
        For illustrative purposes, we shade-out the outputs that do not flow forward, and with $\Sigma$ summarize the aggregations in~\eq{capsule}.
    }
    \label{fig:outline}
    \vspace{-0.5em}
\end{figure*}

\section{Method}
\label{sec:method}
Our network trains on unaligned point clouds as illustrated in \Figure{outline}: we train a network that \textit{decomposes} point clouds into parts, and enforce invariance/equivariance through a Siamese training setup~\cite{keypointnet}.
We then \textit{canonicalize} the point cloud to a learnt frame of reference, and perform \textit{auto-encoding} in this coordinate space.
The losses employed to train $\acne$, $\regressor$, and $\decoder$, will be covered in~\Section{losses}, while the details of their architecture are in~\Section{encoder}.

\paragraph{Decomposition}
In more detail, given a point cloud $\points \in \real^{P{\times}D}$ of $P$ points in $D$ dimensions (in our case $D{=}3$), we perturb it with two random transformations $\mT^a, \mT^b \in \mathbf{SE}(D)$ to produce point clouds $\points^a,\points^b$. 
We then use a shared permutation-equivariant capsule encoder $\acne$ to compute a $K$-fold attention map $\mathbf{A} \in \real^{P{\times}K}$ for $K$ capsules, as well as per-point feature map $\featuremap \in \real^{P{\times}C}$ with $C$ channels:
\begin{equation}
\label{eq:encoder}
\attention, \featuremap = \acne(\points)
\;,
\end{equation}
where we drop the superscript indexing the Siamese branch for simplicity.
From these attention masks, we then compute, for the $k$--th capsule its pose~$\pose_k \in \real^3$ parameterized by its location in 3D space, and the corresponding capsule descriptor $\descriptor_k \in \real^C$:
\begin{align}
\pose_k = \frac{\sum_{p}\bA_{p,k} \points_{p}}{\sum_{p}\bA_{p,k}}
\;,
\qquad
\descriptor_k = \frac{\sum_{p}\bA_{p,k} \featuremap_{p}}{\sum_{p}\bA_{p,k}}
\;.
\label{eq:capsule}
\end{align}
Hence, as long as $\acne$ is invariant w.r.t. rigid transformations of $\points$, the pose $\pose_k$ will be transformation equivariant, and the descriptor $\descriptor_k$ will be transformation invariant.
Note that this simplifies the design (and training) of the encoder $\acne$, which only needs to be invariant, rather than equivariant~\cite{thomas2018tensor,srivastava2019geometric}.

\paragraph{Canonicalization}
Simply enforcing invariance and equivariance with the above framework is not enough to learn 3D representations that are object-centric, as we lack an (unsupervised) mechanism to bring information into a shared~``object-centric'' reference frame.
Furthermore, the ``right'' canonical frame is nothing but a convention, thus we need a mechanism that allows the network to make a \mbox{choice~--~a} choice, however, that must then be consistent across all objects.
For example, a learnt canonical frame where the cockpit of airplanes is consistently positioned along $+z$ is \textit{just as good} as a canonical frame where it is positioned along the $+y$ axis.
To address this, we propose to link the capsule descriptors to the capsule poses in canonical space, that is, we ask that objects with similar appearance to be located in similar Euclidean neighborhoods in canonical space.
We achieve this by regressing canonical capsules poses (i.e. canonical keypoints) $\bar{\pose} \in \real^{K\times3}$ using the descriptors $\descriptor \in \real^{K\times C}$ via a fully connected deep network~$\mathcal{K}$:
\begin{equation}
\bar{\pose} = \regressor\left(\descriptor\right)
\end{equation}
Because fully connected layers are biased towards learning low-frequency representations~\cite{jacot2018neural}, this regressor also acts as a regularizer that enforces semantic locality.

\paragraph{Auto-encoding}
Finally, in the learnt \textit{canonical} frame of reference, to train the capsule descriptors via auto-encoding, we reconstruct the point clouds with per-capsule decoders~$\decoder_k$:
\begin{equation}
\tilde\points = \cup_k \left\{ \decoder_k(\bar\mR\pose_k+\bar\vt, \descriptor_k) \right\}
\;,
\label{eq:recon}
\end{equation}
where $\cup$ denotes the union operator. 
The canonicalizing transformation $\bar\mT=(\bar\mR, \bar\vt)$ can be readily computed by solving a shape-matching problem~\cite{shapematching}, thanks to the property that our capsule poses and regressed keypoints are in \textit{one-to-one correspondence}:
\begin{equation}
\bar\mR, \bar\vt = \argmin_{\mR, \vt} \frac{1}{K} \sum_k \| (\mR \pose_k + \vt) - \bar\pose_k \|_2^2
\;.
\label{eq:procrustes}
\end{equation}
While the reconstruction in \eq{recon} is in canonical frame, note it is trivial to transform the point cloud back to the original coordinate system after reconstruction, as the transformation~$\bar\mT^{-1}$ is available.

\subsection{Losses}
\label{sec:losses}
As common in unsupervised methods, our framework relies on a number of losses that control the different characteristics we seek to obtain in our representation.
Note none of these losses require ground truth labels.
We organize the losses according to the portion of the network they supervise: \textit{decomposition}, \textit{canonicalization}, and \textit{reconstruction}.

\paragraph{Decomposition}
While a transformation invariant encoder architecture should be sufficient to achieve
equivariance/invariance, this does not prevent the encoder from producing trivial solutions/decompositions once trained.
As capsule poses should be \textit{transformation equivariant}, the poses of the two rotation augmentations $\pose_k^a$ and $\pose_k^b$ should only differ by the (known) relative transformation:
\begin{equation}
\loss{equivariance} = \frac{1}{K} \sum_k \| \pose_k^a - (\mT^a)(\mT^b)^{-1}\pose_k^b \|_2^2
\;.
\end{equation}
Conversely, capsule descriptors should be \emph{transformation invariant}, and as the two input points clouds are of the \textit{same} object, the corresponding capsule descriptors $\descriptor$ should be identical:
\begin{equation}
\loss{invariance} = \frac{1}{K} \sum_k \| \descriptor_k^a - \descriptor_k^b \|_2^2
\;.
\end{equation}
We further regularize the capsule decomposition to ensure each of the~$K$ heads roughly represent the same ``amount'' of the input point cloud, hence preventing degenerate (zero attention) capsules. This is achieved by penalizing the attention \textit{variance}:
\begin{equation}
\loss{equilibrium} = \frac{1}{K} \sum_k \| a_k  - \tfrac{1}{K}\Sigma_k a_k \|_2^2
\;,
\label{eq:equilibrium}
\end{equation}
where $a_k=\Sigma_p(\attention_{p,k})$ denotes the total attention exerted by the k-th head on the point cloud.

Finally, to facilitate the training process, we ask for capsules to learn a localized representation of geometry.
We express the spatial extent of a capsule by computing first-order moments of the represented points with respect to the capsule pose $\pose_k$:
\begin{equation}
\loss{localization} = \frac{1}{K} \sum_k \tfrac{1}{a_k} \sum_p \attention_{p,k} \| \pose_k - \points_p \|_2^2 
\;.
\end{equation}

\paragraph{Canonicalization}
To train our canonicalizer $\regressor$, we relate the predicted capsule poses to regressed canonical capsule poses via the \textit{optimal} rigid transformation from~\eq{procrustes}:
\begin{equation}
\loss{canonical} = \frac{1}{K} \sum_k \| (\bar\mR \pose_k + \bar\vt) - \bar\pose_k \|_2^2
\;.
\label{sec:procrustes}
\end{equation}
Recall that $\bar\mR$ and $\bar\mT$ are obtained through a 
differentiable process.
Thus, this loss is forcing the aggregated pose $\pose_k$ to agree with the one that goes through the regression path, $\bar\pose_k$.
Now,
since $\bar\pose_k$ is regressed solely from the set of capsule descriptors, similar shapes will result in similar canonical keypoints, and the coordinate system of $\bar\pose_k$ is one that employs Euclidean space to encode semantics.

\paragraph{Reconstruction}
To learn canonical capsule descriptors in an unsupervised fashion, we rely on an auto-encoding task.
We train the decoders $\{\mathcal{D}_k\}$ by minimizing the \textit{Chamfer Distance}~(CD) between the (canonicalized) input point cloud and the reconstructed one, as in~\cite{yang2018foldingnet,groueix2018papier}: 
\begin{equation}
\loss{recon} = \text{CD} \left( \bar\mR \points + \bar\vt, \: \tilde\points \right)
\;.
\end{equation}

\subsection{Network Architectures}
We briefly summarize our implementation details, including the network architecture; for further details, please refer to the \SupplementaryMaterial.
\label{sec:arch}

\paragraph{Encoder -- $\acne$}
\label{sec:encoder}
Our architecture is based on the one suggested in~\cite{sun2020acne}: a pointnet-like architecture with residual connections and attentive context normalization.
We utilize Batch Normalization~\cite{Ioffe15} instead of the Group Normalization~\cite{Wu18a}, which trained faster in our experiments. 
We further extend their method to have \textit{multiple} attention maps, where each attention map corresponds to a capsule.

\paragraph{Decoder -- $\decoder$}
\label{sec:decoder}
The decoder from \eq{recon} operates on a per-capsule basis. 
Our decoder architecture is similar to AtlasNetV2~\cite{deprelle2019learning} (with trainable grids). 
The difference is that we translate the per-capsule decoded point cloud by the corresponding capsule pose.

\paragraph{Regressor -- $\regressor$}
\label{sec:regressor}
We 
concatenate the descriptors and apply a series of fully connected layers with ReLU activation to regress the $P$ capsule locations.
At the output layer we use a linear activation and subtract the mean of the outputs to make our regressed locations zero-centered in the canonical frame.

\paragraph{Canonicalizing the descriptors}
As our descriptors are only \textit{approximately} rotation invariant (via $\loss{invariance}$), we found it useful to re-extract the capsule descriptors $\descriptor_k$ after canonicalization.
Specifically, we compute $\bar\featuremap$ with the same encoder setup, but with $\bar\points {=} \bar\mR\points {+} \bar\mT$ instead of $\points$ and use it to compute $\bar\descriptor_k$; we validate this empirically in the \SupplementaryMaterial.

\section{Results}
We first discuss the experimental setup, and then validate our method on a \textit{variety} of tasks: auto-encoding, canonicalization, and unsupervised classification. 
While the task differs, our learning process remains the \textit{same}: we learn capsules by reconstructing objects in a learnt canonical frame.
We also provide an ablation study, which is expanded in detail in the \SupplementaryMaterial, where we provide additional qualitative results.

\subsection{Experimental setup}
\label{sec:setup}
To evaluate our method, we rely on the ShapeNet (Core) dataset~\cite{shapenet}\footnote{Available to researchers for non-commercial research and educational use.}.
We follow the category choices from AtlasNetV2~\cite{deprelle2019learning}, using the airplane and chair classes for \textit{single-category} experiments, while for \textit{multi-category} experiments we use all $13$ classes: airplane, bench, cabinet, car, chair, monitor, lamp, speaker, firearm, couch, table, cellphone, and watercraft.
To make our results most compatible with those reported in the literature, we also use the same splits as in AtlasNetV2~\cite{deprelle2019learning}: $31747$ shapes in the train, and $7943$ shapes in the test set.
Unless otherwise noted, we randomly sample $1024$ points from the object surface for each shape to create our 3D point clouds.

\paragraph{De-canonicalizing the dataset}
As discussed in the introduction, ShapeNet (Core) contains substantial inductive bias in the form of consistent semantic alignment.
To remove this bias, we create random \SE{3} transformations, and apply them to each point cloud.
We first generate uniformly sampled random rotations, and add uniformly sampled random translations within the range $[-0.2, 0.2]$, where the bounding volume of the shape ranges in $[-1,+1]$.
Note the relatively limited translation range is chosen to give state-of-the-art methods a \textit{chance} to compete with our solution.
We then use the relative transformation between the point clouds extracted from this ground-truth transformation to evaluate our methods.
We refer to this unaligned version of the ShapeNet Core dataset as the \textit{unaligned} setup, and using the vanilla ShapeNet Core dataset as the \textit{aligned} setup.
For the \textit{aligned} setup, as there is no need for equivariance adaptation, we simply train our method without the random transformations, and so $\loss{equivariance}$ and $\loss{invariance}$ are not used.
This setup is to simply demonstrate how Canonical Capsules would perform in the presence of a dataset bias.

We emphasize here that a proper generation of random rotation is important.
While some existing works have generated them by uniformly sampling the degrees of freedom of an Euler-angle representation, this is known to be an incorrect way to sample random rotations~\cite{shoemake1992uniform}, leading to biases in the generated dataset; see the \SupplementaryMaterial.

\paragraph{Implementation details}
For all our experiments we use the Adam optimizer~\cite{Kingma15} with an initial learning rate of $0.001$ and decay rate of $0.1$.
We train for $325$ epochs for the $\it aligned$ setup to match the AtlasNetV2~\cite{deprelle2019learning} original setup.
For the $\it unaligned$ setting, as the problem is harder, we train for a longer number of $450$ epochs.
We use a batch size of 16. 
The training rate is {$\sim$2.5 iters/sec}.
We train each model on a single NVidia V100 GPU.
Unless stated otherwise, we use $k{=}10$ capsules and capsule descriptors of dimension $C{=}128$.
We train three models with our method: two that are single-category (\ie,~for airplane and chairs), and one that is multi-category (\ie,~all 13 classes).
To set the weights for each loss term, we rely on the reconstruction performance (CD) in the training set. We set weights to be one for all terms except for $\loss{equivariance}$ (5) and $\loss{equilibrium}$ ($10^{-3}$). 
In the aligned case, because $\loss{equivariance}$ and $\loss{invariance}$ are not needed (always zero), we reduce the weights for the other decomposition losses by $10^{3}$; $\loss{localization}$ to $10^{-3}$ and $\loss{equilibrium}$ to $10^{-6}$.

\begin{wrapfigure}[11]{r}{0.45\linewidth}
    \vspace{-1.1em}
    \captionof{table}{
        \textbf{Auto-encoding / quantitative --} 
        Performance in terms of Chamfer distance with 1024 points per point cloud -- metric is multiplied by $10^3$ as in \cite{deprelle2019learning}.
    } %
    \vspace{-.7em}
    \begin{center}
        \resizebox{\linewidth}{!}{
            \begin{tabular}{llccc}
                \toprule
                & Method & Airplane  & Chair  & Multi\\
                \midrule
                \multirow{3}{*}{\rotatebox[origin=c]{90}{\footnotesize Aligned}}
                & \capsnet & 1.94    & 3.30   & 2.49 \\ 
                & \atlasnet & 1.28  & 2.36   & 2.14  \\
                & Our method   & $\mathbf{0.96}$ & $\mathbf{1.99}$     & $\mathbf{1.76}$ \\
                
                \midrule
                \multirow{4}{*}{\rotatebox[origin=c]{90}{\footnotesize Unaligned}} &
                \capsnet & 5.58   & 7.57 & $4.66$   \\ 
                & \atlasnet & 2.80  & 3.98 & 3.08   \\
                & \atlasnet w/ STN& 1.90  & 2.99 & 2.60   \\
                & Our method & \textbf{1.11} & \textbf{2.58} & \textbf{2.22} \\
                \bottomrule
            \end{tabular}
        } %
    \end{center}
    \label{tbl:both_recons}
\end{wrapfigure}
 
\subsection{Auto-encoding -- \Figure{qualitative} and \Table{both_recons}}
\label{sec:reconstruction}
We evaluate the performance of our method for the task that was used to train the network -- reconstruction / auto-encoding -- against three baselines (trained in both single-category and multi-category variants):
\CIRCLE{1} \capsnet, an auto-encoder for 3D point clouds that utilize a capsule architecture;
\CIRCLE{2} \atlasnet, a state-of-the-art auto-encoder which utilizes a multi-head patch-based decoder;
\CIRCLE{3} \atlasnet with a spatial-transformer network (STN) aligner from PointNet~\cite{qi2017pointnet}, a baseline with canonicalization.
We do not compare against~\cite{srivastava2019geometric}, as unfortunately source code is not publicly available. 

\newcommand{\footpoints}{Results in this table differ slightly from what is reported in the original papers as we use 1024 points to speed-up experiments throughout our paper. However, in the \SupplementaryMaterial the same trend holds regardless of the number of points, and match with what is reported in the original papers with 2500 points.}

\paragraph{Quantitative analysis -- \Table{both_recons}}
We achieve state-of-the-art performance in both the {\it aligned} and {\it unaligned} settings.
The wider margin in the \textit{unaligned} setup indicates tackling this more challenging scenario damages the performance of \atlasnet and \capsnet much more than our method\footnote{\footpoints}.
We also include a variant of AtlasNetV2 for which a STN (Spatial Transformer Network) is used to pre-align the point clouds~\cite{qi2017pointnet}, demonstrating how the simplest form of pre-aligner/canonicalizer is not sufficient.

\paragraph{Qualitative analysis -- \Figure{qualitative}}
We illustrate our decomposition-based reconstruction of 3D point clouds, as well as the reconstructions of \capsnet and \atlasnet.
As shown, even in the $\it unaligned$ setup, our method is able to provide \textit{semantically consistent} capsule decompositions -- \eg the wings of the airplane have consistent colours, and when aligned in the canonical frame, the different airplane instances are well-aligned.
Compared to \atlasnet and \capsnet, the reconstruction quality is also visibly improved: we 
better 
preserve details along the engines of the airplane, or the thin structures of the bench; note also that the decompositions are semantically consistent in our examples.
Results are better appreciated in our \SupplementaryMaterial, where we visualize the performance as we continuously traverse \SE{3}.

\def\capshift{-5}
\def \qualfigwidth {0.162}
\def \qualcapwidth {0.13}
\begin{figure*}
    \begin{center}
        \begin{overpic}
            [width=\linewidth]
            {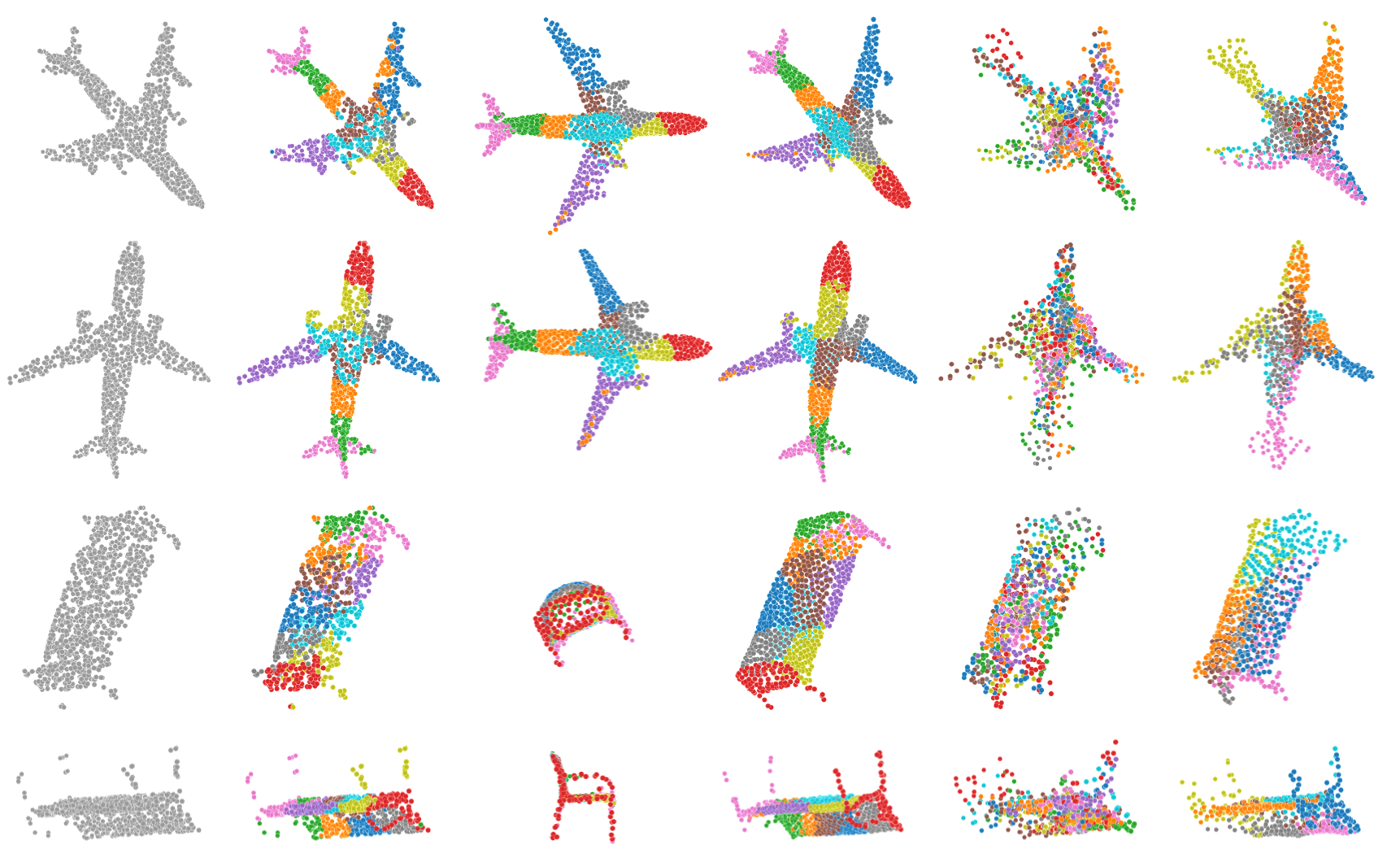}
            \put(0,\capshift){\parbox{\qualcapwidth\linewidth}{\centering \scriptsize Input}}
            \put(16,\capshift){\parbox{\qualcapwidth\linewidth}{\centering \scriptsize \textit{Our} capsule decomposition}}
            \put(34,\capshift){\parbox{\qualcapwidth\linewidth}{\centering \scriptsize \textit{Our} reconstruction in canonical frame}}
            \put(51,\capshift){\parbox{\qualcapwidth\linewidth}{\centering \scriptsize \textit{Our} reconstruction in input frame}}
            \put(66,\capshift){\parbox{\qualcapwidth\linewidth}{\centering \scriptsize \capsnet reconstruction}}
            \put(84,\capshift){\parbox{\qualcapwidth\linewidth}{\centering \scriptsize \atlasnet reconstruction}}
        \end{overpic}
        \vspace{1em}
    \end{center}
\caption{
\textbf{Auto-encoding / qualitative -- }
Example decomposition and reconstruction results setup using Canonical Capsules on several $\it unaligned$ point cloud instances from the test set.
We color each Canonical Capsule with a unique colour, and similarly color ``patches'' from the reconstruction heads of \capsnet and \atlasnet.
Canonical Capsules provide semantically consistent decomposition that is aligned in canonical frame, leading to improved reconstruction quality.
}
\label{fig:qualitative}
\end{figure*}

\subsection{Canonicalization -- \Table{canonicalization}}
\label{sec:canonicalization}
We compare against three baselines:
\CIRCLE{1} Deep Closest Points~\cite{wang2019deep}, a deep learning-based \textit{pairwise} point cloud registration method;
\CIRCLE{2} DeepGMR~\cite{yuan2020deepgmr} a state-of-the-art \textit{pairwise} registration method that decomposes clouds into Gaussian mixtures and utilizes Rigorously Rotation-Invariant (RRI) features~\cite{chen2019clusternet};
\CIRCLE{3} Compass~\cite{spezialetti2020learning} a concurrent work on learnt alignment/canonicalization.
For all compared methods we use the official implementation.
For DeepGMR we use both RRI and the typical XYZ coordinates as input.
We also try our method with the RRI features, following DeepGMR's training protocol and train for 100 epochs.

\paragraph{Metrics}
To evaluate the canonicalization performance, we look into the stability of the canonicalization -- the shakiness shown in the videos in our \SupplementaryMaterial -- represented as the mean standard deviation of the rotations (mStd):
\begin{equation}
\text{mStd} = \frac{1}{n}\sum\nolimits_{i=1}^{n}\sqrt{\frac{1}{m}\sum\nolimits_{j=1}^m (\angle(\mR^{ij}, {\mR}_{mean}^i))^2 }
\;,
\end{equation}
where $\angle$ is the angular distance between two rotation matrices~\cite{gojcic2020learning, yew2020rpm, yuan2020deepgmr},
$\mR^{ij}$ is the rotation matrix of the $j$-th instance of the $i$-th object in canonical frame, and ${\mR}_{mean}^i$ is the mean rotation~\cite{hartley2013rotation} of the $i$-th object.
Note that with mStd we measure the stability of canonicalization with respect to rotations to accommodate for methods that do not deal with translation~\cite{spezialetti2020learning}.
To allow for comparisons with pairwise registration methods, we also measure performance in terms of the RMSE metric~\cite{yuan2020deepgmr, zhou2016fast}.

\paragraph{Quantitative analysis -- \Table{canonicalization}}
Compared to \compass, our method provides improved stability in canonicalization.
This also provides an advantage in pairwise registration, delivering state-of-the-art results when XYZ-coordinates are used.
Note that while \textit{pairwise} methods can align two sets of given point clouds, creating a canonical frame that simultaneously registers \textit{all} point clouds is a non-trivial extension to the problem.

When RRI is used as input, our method is on par with \deepgmr, up to a level where registration is near perfect -- alignment differences when errors are in the $10^{-4}$ ballpark are indiscernible.
We note that the performance of Deep Closest Points~\cite{wang2019deep} is not as good as reported in the original paper, as we uniformly draw rotations from \SO{3}. 
When a sub-portion of \SO{3} is used, \eg a quarter of what we are using, DCP performs relatively well (0.008 in the multi-class experiment).
While curriculum learning could be used to enhance the performance of DCP, our technique does not need to rely on these more complex training techniques.

We further note that, while RRI delivers good registration performance, using RRI features cause the learnt canonicalization to fail -- training becomes \textit{unstable}.
This hints that RRI features may be throwing away too much information to achieve transformation invariance.
Our method using raw XYZ coordinates as input, on the other hand, provides comparable registration performance, and is able to do significantly more than just registration (\ie classification, reconstruction).

\begin{table}
\caption{
\textbf{Canonicalization --}
Quantitative evaluation for canonicalization, where we highlight significantly better performance than the concurrent work Compass~\cite{spezialetti2020learning}.
While pairwise registration is not our main objective, the \textit{global} alignment frame created by our method still allows for effective registration (on par, or better) than the state-of-the-art.
}
    \begin{center}
        \setlength{\tabcolsep}{12pt}
        \resizebox{\linewidth}{!}{
        \small
            \begin{tabular}{llcccccc}
                \toprule
                                                        & & \multicolumn{3}{c}{Canonicalization (mStd)~$\downarrow$} & \multicolumn{3}{c}{Pairwise registration (RMSE)~$\downarrow$}   \\
                                                        \cmidrule(r){3-5}
                                                        \cmidrule{6-8}
                                                        
                                                        &Method& Airplane        & Chair           & Multi        & Airplane & Chair & Multi   \\
                \midrule
                \multirow{4}{*}{\rotatebox[origin=c]{90}{\scriptsize XYZ-coord.}}
                & Deep Closest Points~\cite{wang2019deep} & -- & -- & -- & 0.318           & 0.160           & 0.131           \\
                & \deepgmr & -- & -- & -- & 0.079           & 0.082           & 0.077           \\
                & \compass &19.105& 19.508 & 51.811&0.412            &0.482            & 0.515           \\
                & Our method & \textbf{8.278} &\textbf{5.730} & \textbf{21.210}& \textbf{0.022}  & \textbf{0.027}  & \textbf{0.074}  \\
                \midrule
                \multirow{2}{*}{\rotatebox[origin=c]{90}{\scriptsize RRI}}
                
                & DeepGMR~\cite{yuan2020deepgmr}  & -- & -- & -- & \textbf{0.0001} & \textbf{0.0001} & \textbf{0.0001} \\
                & Our method & {\it (unstable)} & {\it (unstable)} & {\it (unstable)} & 0.0006          & 0.0009          & 0.0016          \\
                \bottomrule
            \end{tabular}
        } %
    \end{center}
\label{tbl:canonicalization}
\vspace{-1em}
\end{table}

\subsection{Unsupervised classification -- \Table{both_class}}
\label{sec:classification}
Beyond reconstruction and canonicalization,  we evaluate the usefulness of our method via a classification task that is not related in \textit{any way} to the losses used for training.
We compute the features from the auto-encoding methods from \Section{reconstruction} against those from our method~(where we build features by combining pose with descriptors) to perform $13$-way classification with two different techniques:
\begin{wrapfigure}[10]{r}{0.45\linewidth}
    \vspace{0em}
    \captionof{table}{
        \textbf{Classification} -- Top-$1$~accuracy(\%)
    }
    \vspace{-1em}
    \begin{center}
        \resizebox{\linewidth}{!}{
            \begin{tabular}{llcc}
                \toprule
                & Method & SVM & K-Means \\
                \midrule
                \multirow{3}{*}{\rotatebox[origin=c]{90}{\footnotesize Aligned}} 
                & \capsnet & 93.81  & 65.87\\
                & \atlasnet & 94.07  & 61.66   \\
                & Our method & $\mathbf{94.21}$ & $\mathbf{69.82}$\\
                \midrule
                \multirow{4}{*}{\rotatebox[origin=c]{90}{\footnotesize Unaligned}} 
                & \capsnet & 71.13  & 14.59\\
                & \atlasnet & 64.85 & 17.12\\
                & \atlasnet w/ STN& 78.55& {20.03}\\
                & Our method & \textbf{87.33} & \textbf{43.04}\\
                \bottomrule
            \end{tabular}
        } %
    \end{center}
    \label{tbl:both_class}
\end{wrapfigure}
\vspace{-1em}
\begin{itemize}[leftmargin=*]
    \setlength\itemsep{0em}
    \item We train a \textit{supervised} linear Support Vector Machine~(SVM) on the extracted features~\cite[Ch.~7]{prml};
    \item We perform \textit{unsupervised} K-Means clustering~\cite[Ch.~9]{prml} and then label each cluster via bipartite matching with the actual labels through the Hungarian algorithm.
\end{itemize}
\vspace{-.5em}
Note the former provides an upper bound for unsupervised classification, while better performance on the latter implies that the learnt features are able to separate the classes into clusters that are compact~(in an Euclidean sense).

\paragraph{Analysis of results -- SVM}
Note how our method provides best results in all cases, and when the dataset is not unaligned the difference is significant.
This shows that, while 3D-PointCapsNet and AtlasNetV2 (with and without STN) are able to somewhat auto-encode point clouds in the $\it unaligned$ setup, what they learn does not translate well to classification.
However, the features learned with Canonical Capsules are more related to the semantics of the object, which helps classification.

\paragraph{Analysis of results -- K-Means}
The performance gap becomes wider when K-Means is used -- even in the $\it aligned$ case.
This could mean that the features extracted by Canonical Capsules are better suited for other unsupervised tasks, having a feature space that is close to being Euclidean in terms of semantics.
The difference is striking in the $\it unaligned$ setup.
We argue that these results emphasize the importance of the capsule framework -- jointly learning the invariances and equivariances in the data -- is cardinal to unsupervised learning~\cite{hinton1981parallel,hinton1985shape}.

\subsection{Ablation study}
\label{sec:ablation}
To make the computational cost manageable, we perform all ablations with the \textit{airplane} category (the category with most instances), and in the $\it unaligned$ setup (unless otherwise noted).
Please also see \SupplementaryMaterial for more ablation studies.

\begin{table}
    \begin{minipage}{0.52\linewidth}
    \captionof{table}{
        \textbf{Effect of losses -- }
        Reconstruction performance, and canonicalization performance when loss terms are removed.
    }
    \begin{center}
        \resizebox{\linewidth}{!}{
            \begin{tabular}{c|cccccc}
                \toprule
                  & Full          & $\lnot\loss{invar}$ & $\lnot\loss{equiv}$ & $\lnot\loss{canonical}$ & $\lnot\loss{localization}$ & $\lnot\loss{equilibrium}$ \\
                \midrule
                CD & \textbf{1.11} & 1.12                & 1.16                    & 1.12                & 1.44                       & 1.60                      \\
                Std & \textbf{8.278} &{9.983}& {110.174} & {8.421} &{113.204}&{92.970} \\
                \bottomrule
            \end{tabular}
        } %
    \end{center}
    \label{tbl:loss_terms}
\end{minipage}

    \hfill
    \begin{minipage}{0.42\linewidth}
\captionof{table}{
    \textbf{Backbone --}
    Auto-encoding performance (Chamfer distance) when we use various permutation-invariant backbones.
}
\begin{center}
    \resizebox{\linewidth}{!}{
        \begin{tabular}{c|cccc}
            \toprule
                & PointNet & PointNet++ & DGCNN & ACNe  \\
            \midrule
            CD & 1.28 & 1.34 & 1.21 & \textbf{1.11}\\
            \bottomrule
        \end{tabular}
    } %
\end{center}
\label{tbl:ablation_backbone}
\end{minipage}
 
    \vspace{-1em}
\end{table}

\paragraph{Losses -- \Table{loss_terms}}
We analyze the importance of each loss term, with the exception of $\loss{recon}$ which is necessary for training.
All losses beneficially contribute to reconstruction performance, but note how $\loss{equiv}$, $\loss{localization}$ and $\loss{equilibrium}$ affect it to a larger degree.
By considering our canonicalization metric, we can motivate this outcome by observing that the method fails to perform canonicalization when these losses are not employed~(i.e.~training collapses).

\paragraph{Encoder architecture -- \Table{ablation_backbone}}
Our method can be used with \textit{any} backbone, as our main contribution lies in the self-supervised canonicalization architecture.
For completeness, we explore variants of our method using different back-bones:
PointNet~\cite{qi2017pointnet}, PointNet++~\cite{qi2017pointnetplusplus}, DGCNN~\cite{wang2019dynamic}, and ACNe~\cite{sun2020acne}.
Among these, the ACNe backbone performs best; note that all the variants in~\Table{ablation_backbone}, \textit{regardless} of backbone choice, significantly outperforms all other methods reported in~\Table{both_recons}.

\section{Conclusions}
\label{sec:conclusion}

In this paper, we provide a self-supervised framework to train \textit{primary} capsule decompositions for 3D point clouds.
We rely on a Siamese setup that allows self-supervision and auto-encoding in canonical space, circumventing the customary need to train on pre-aligned datasets.
Despite being trained in a self-supervised fashion, our representation achieves state-of-the-art performance across auto-encoding, canonicalization and classification tasks.
These results are made possible by allowing the network to learn a canonical frame of reference.
We interpret this result as giving our neural networks a mechanism to construct a ``mental picture'' of a given 3D object -- so that downstream tasks are executed within an object-centric coordinate frame.

\paragraph{Future work}
As many objects have natural symmetries that we do not consider at the moment~\cite{deprelle2019learning}, providing our canonicalizer a way to encode such a prior is likely to further improve the representation.
We perform decomposition at a single level, and it would be interesting to investigate how to effectively engineer multi-level decompositions~\cite{conshier_sig13};
one way could be to over-decompose the input in a redundant fashion (with large $K$), and use a downstream layers that ``selects'' the decomposition heads to be used~\cite{chen2020bsp}.
We would also like to extend our results to more ``in-the-wild'' 3D computer vision and understand whether learning object-centric representations is possible when \textit{incomplete} (\ie, single view~\cite{park2019deepsdf} or occluded) data is given in input, when an entire scene with potentially \textit{multiple} objects is given~\cite{song2016ssc}, or where our measurement of the 3D world is a single 2D image~\cite{keypointnet}, or by exploiting the persistence of objects in video~\cite{sabour2021flowcaps}.

\paragraph{Broader impact}
While our work is exclusively on 3D shapes, and thus not immediately subject to any societal concerns, it enhances how Artificial Intelligence (AI) can understand and model 3D geometry.
Thus, similar to how image recognition could be misused, one should be careful when extending the use of our method.
In addition, while not subject to how the data itself is aligned, the learnt canonical frame of our method is still data-driven, thus subject to any data collection biases that may exist -- canonicalization will favour shapes that appear more often. 
This should also be taken into account with care to prevent any biased decisions when utilizing our method within a decision making AI platform.

\section*{Acknowledgements}
This work was supported by the Natural Sciences and Engineering Research Council of Canada (NSERC) Discovery Grant, NSERC Collaborative Research and Development Grant, Google, Compute Canada, and Advanced Research Computing at the University of British Columbia.

{
    \small
    \bibliographystyle{ieee}
    \bibliography{macros,main}

\begin{thebibliography}{10}\itemsep=-1pt

\bibitem{prml}
Christopher~M Bishop.
\newblock {\em {Pattern Recognition and Machine Learning}}.
\newblock springer, 2006.

\bibitem{chabra2020deep}
Rohan Chabra, Jan~Eric Lenssen, Eddy Ilg, Tanner Schmidt, Julian Straub, Steven
  Lovegrove, and Richard Newcombe.
\newblock {Deep Local Shapes: Learning Local SDF Priors for Detailed 3D
  Reconstruction}.
\newblock In {\em European Conference on Computer Vision}, 2020.

\bibitem{shapenet}
Angel~X Chang, Thomas Funkhouser, Leonidas Guibas, Pat Hanrahan, Qixing Huang,
  Zimo Li, Silvio Savarese, Manolis Savva, Shuran Song, Hao Su, et~al.
\newblock {Shapenet: An Information-Rich 3D Model Repository}.
\newblock {\em arXiv Preprint}, 2015.

\bibitem{chen2019clusternet}
Chao Chen, Guanbin Li, Ruijia Xu, Tianshui Chen, Meng Wang, and Liang Lin.
\newblock {Clusternet: Deep Hierarchical Cluster Network with Rigorously
  Rotation-Invariant Representation for Point Cloud Analysis}.
\newblock In {\em Conference on Computer Vision and Pattern Recognition}, 2019.

\bibitem{chen2020simple}
Ting Chen, Simon Kornblith, Mohammad Norouzi, and Geoffrey Hinton.
\newblock {A Simple Framework for Contrastive Learning of Visual
  Representations}.
\newblock {\em International Conference on Machine Learning}, 2020.

\bibitem{chen2020bsp}
Zhiqin Chen, Andrea Tagliasacchi, and Hao Zhang.
\newblock {Bsp-net: Generating compact meshes via binary space partitioning}.
\newblock In {\em Conference on Computer Vision and Pattern Recognition}, 2020.

\bibitem{chen2019learning}
Zhiqin Chen and Hao Zhang.
\newblock {Learning Implicit Fields for Generative Shape Modeling}.
\newblock In {\em Conference on Computer Vision and Pattern Recognition}, 2019.

\bibitem{deng2019neural}
Boyang Deng, JP Lewis, Timothy Jeruzalski, Gerard Pons-Moll, Geoffrey Hinton,
  Mohammad Norouzi, and Andrea Tagliasacchi.
\newblock {NASA: Neural Articulated Shape Approximation}.
\newblock In {\em European Conference on Computer Vision}, 2020.

\bibitem{deng2021vn}
Congyue Deng, Or Litany, Yueqi Duan, Adrien Poulenard, Andrea Tagliasacchi, and
  Leonidas Guibas.
\newblock Vector neurons: a general framework for so(3)-equivariant networks,
  2021.

\bibitem{deng2020better}
Zhantao Deng, Jan Bedna{\v{r}}{\'\i}k, Mathieu Salzmann, and Pascal Fua.
\newblock {Better Patch Stitching for Parametric Surface Reconstruction}.
\newblock {\em arXiv Preprint}, 2020.

\bibitem{deng2020cvxnet}
{Deng, Boyang and Genova, Kyle and Yazdani, Soroosh and Bouaziz, Sofien and
  Hinton, Geoffrey and Tagliasacchi, Andrea}.
\newblock {CvxNet: Learnable Convex Decomposition}.
\newblock In {\em Conference on Computer Vision and Pattern Recognition}, 2020.

\bibitem{deprelle2019learning}
Theo Deprelle, Thibault Groueix, Matthew Fisher, Vladimir Kim, Bryan Russell,
  and Mathieu Aubry.
\newblock {Learning Elementary Structures for 3D Shape Generation and
  Matching}.
\newblock In {\em Advances in Neural Information Processing Systems}, 2019.

\bibitem{fan2017point}
Haoqiang Fan, Hao Su, and Leonidas~J Guibas.
\newblock {A Point Set Generation Network for 3D Object Reconstruction from a
  Single Image}.
\newblock In {\em Conference on Computer Vision and Pattern Recognition}, 2017.

\bibitem{dpm}
Pedro Felzenszwalb, David McAllester, and Deva Ramanan.
\newblock {A Discriminatively Trained, Multiscale, Deformable Part Model}.
\newblock In {\em Conference on Computer Vision and Pattern Recognition}, 2008.

\bibitem{fernandez2020unsupervised}
Clara Fernandez-Labrador, Ajad Chhatkuli, Danda~Pani Paudel, Jose~J Guerrero,
  C{\'e}dric Demonceaux, and Luc Van~Gool.
\newblock {Unsupervised Learning of Category-Specific Symmetric 3D Keypoints
  from Point Sets}.
\newblock In {\em European Conference on Computer Vision}, 2020.

\bibitem{genova2019deep}
Kyle Genova, Forrester Cole, Avneesh Sud, Aaron Sarna, and Thomas Funkhouser.
\newblock {Deep Structured Implicit Functions}.
\newblock In {\em Conference on Computer Vision and Pattern Recognition}, 2020.

\bibitem{genova2019learning}
Kyle Genova, Forrester Cole, Daniel Vlasic, Aaron Sarna, William~T Freeman, and
  Thomas Funkhouser.
\newblock {Learning Shape Templates with Structured Implicit Functions}.
\newblock In {\em International Conference on Computer Vision}, 2019.

\bibitem{gojcic2020learning}
Zan Gojcic, Caifa Zhou, Jan~D Wegner, Leonidas~J Guibas, and Tolga Birdal.
\newblock {Learning multiview 3D point cloud registration}.
\newblock In {\em Conference on Computer Vision and Pattern Recognition}, 2020.

\bibitem{groueix2018papier}
Thibault Groueix, Matthew Fisher, Vladimir~G Kim, Bryan~C Russell, and Mathieu
  Aubry.
\newblock {A Papier-M{\^a}ch{\'e} Approach to Learning 3D Surface Generation}.
\newblock In {\em Conference on Computer Vision and Pattern Recognition}, 2018.

\bibitem{gu2020weakly}
Jiayuan Gu, Wei-Chiu Ma, Sivabalan Manivasagam, Wenyuan Zeng, Zihao Wang, Yuwen
  Xiong, Hao Su, and Raquel Urtasun.
\newblock {Weakly-Supervised 3D Shape Completion in the Wild}.
\newblock In {\em European Conference on Computer Vision}, 2020.

\bibitem{hartley2013rotation}
Richard Hartley, Jochen Trumpf, Yuchao Dai, and Hongdong Li.
\newblock {Rotation averaging}.
\newblock {\em International Journal of Computer Vision}, 2013.

\bibitem{he2016deep}
Kaiming He, Xiangyu Zhang, Shaoqing Ren, and Jian Sun.
\newblock {Deep Residual Learning for Image Recognition}.
\newblock In {\em Conference on Computer Vision and Pattern Recognition}, 2016.

\bibitem{hinton2011transforming}
Geoffrey~E Hinton, Alex Krizhevsky, and Sida~D Wang.
\newblock {Transforming Auto-Encoders}.
\newblock In {\em International Conference on Artificial Neural Networks},
  2011.

\bibitem{hinton1985shape}
Geoffrey~E Hinton and Kevin~J Lang.
\newblock {Shape Recognition and Illusory Conjunctions}.
\newblock In {\em International Joint Conference on Artificial Intelligence},
  1985.

\bibitem{hinton1981parallel}
Geoffrey~F Hinton.
\newblock {A Parallel Computation that Assigns Canonical Object-based Frames of
  Reference}.
\newblock In {\em International Joint Conference on Artificial Intelligence},
  1981.

\bibitem{Ioffe15}
Sergey Ioffe and Christian Szegedy.
\newblock {Batch Normalization: Accelerating Deep Network Training by Reducing
  Internal Covariate Shift}.
\newblock In {\em International Conference on Machine Learning}, 2015.

\bibitem{jacot2018neural}
Arthur Jacot, Franck Gabriel, and Cl{\'e}ment Hongler.
\newblock {Neural Tangent Kernel: Convergence and Generalization in Neural
  Networks}.
\newblock In {\em Advances in Neural Information Processing Systems}, 2018.

\bibitem{kar2017learning}
Abhishek Kar, Christian H{\"a}ne, and Jitendra Malik.
\newblock {Learning a Multi-View Stereo Machine}.
\newblock In {\em Advances in Neural Information Processing Systems}, 2017.

\bibitem{Kingma15}
D.P. Kingma and J. Ba.
\newblock {Adam: {A} Method for Stochastic Optimisation}.
\newblock In {\em International Conference on Learning Representations}, 2015.

\bibitem{kosiorek2019stacked}
Adam Kosiorek, Sara Sabour, Yee~Whye Teh, and Geoffrey~E Hinton.
\newblock {Stacked Capsule Autoencoders}.
\newblock In {\em Advances in Neural Information Processing Systems}, 2019.

\bibitem{litany2018deformable}
Or Litany, Alex Bronstein, Michael Bronstein, and Ameesh Makadia.
\newblock {Deformable Shape Completion with Graph Convolutional Autoencoders}.
\newblock In {\em Conference on Computer Vision and Pattern Recognition}, 2018.

\bibitem{sift}
David~G Lowe.
\newblock {Distinctive Image Features from Scale-Invariant Keypoints}.
\newblock {\em International Journal of Computer Vision}, 60:91--110, 2004.

\bibitem{mescheder2019occupancy}
Lars Mescheder, Michael Oechsle, Michael Niemeyer, Sebastian Nowozin, and
  Andreas Geiger.
\newblock {Occupancy Networks: Learning 3D Reconstruction in Function Space}.
\newblock In {\em Conference on Computer Vision and Pattern Recognition}, 2019.

\bibitem{novotny2019c3dpo}
David Novotny, Nikhila Ravi, Benjamin Graham, Natalia Neverova, and Andrea
  Vedaldi.
\newblock {C3dpo: Canonical 3d pose networks for non-rigid structure from
  motion}.
\newblock In {\em Conference on Computer Vision and Pattern Recognition}, 2019.

\bibitem{park2019deepsdf}
Jeong~Joon Park, Peter Florence, Julian Straub, Richard Newcombe, and Steven
  Lovegrove.
\newblock {DeepSDF: Learning Continuous Signed Distance Functions for Shape
  Representation}.
\newblock In {\em Conference on Computer Vision and Pattern Recognition}, 2019.

\bibitem{qi2017pointnet}
Charles~R Qi, Hao Su, Kaichun Mo, and Leonidas~J Guibas.
\newblock {Pointnet: Deep Learning on Point Sets for 3D Classification and
  Segmentation}.
\newblock In {\em Conference on Computer Vision and Pattern Recognition}, 2017.

\bibitem{qi2017pointnetplusplus}
Charles~R Qi, Li Yi, Hao Su, and Leonidas~J Guibas.
\newblock {PointNet++: Deep Hierarchical Feature Learning on Point Sets in a
  Metric Space}.
\newblock {\em Advances in Neural Information Processing Systems}, 2017.

\bibitem{yolo}
Joseph Redmon, Santosh Divvala, Ross Girshick, and Ali Farhadi.
\newblock {You Only Look Once: Unified, Real-Time Object Detection}.
\newblock In {\em Conference on Computer Vision and Pattern Recognition}, 2016.

\bibitem{rempe2020caspr}
Davis Rempe, Tolga Birdal, Yongheng Zhao, Zan Gojcic, Srinath Sridhar, and
  Leonidas~J. Guibas.
\newblock {CaSPR: Learning Canonical Spatiotemporal Point Cloud
  Representations}.
\newblock In {\em Advances in Neural Information Processing Systems}, 2020.

\bibitem{sabour2021flowcaps}
Sara Sabour, Andrea Tagliasacchi, Soroosh Yazdani, Geoffrey~E. Hinton, and
  David~J. Fleet.
\newblock Unsupervised part representation by flow capsules, 2021.

\bibitem{shoemake1992uniform}
Ken Shoemake.
\newblock Uniform random rotations.
\newblock In {\em Graphics Gems III (IBM Version)}, pages 124--132. Elsevier,
  1992.

\bibitem{simonyan2014very}
Karen Simonyan and Andrew Zisserman.
\newblock {Very Deep Convolutional Networks for Large-Scale Image Recognition}.
\newblock {\em International Conference on Learning Representations}, 2015.

\bibitem{song2016ssc}
Shuran Song, Fisher Yu, Andy Zeng, Angel~X Chang, Manolis Savva, and Thomas
  Funkhouser.
\newblock Semantic scene completion from a single depth image.
\newblock {\em Proceedings of 30th IEEE Conference on Computer Vision and
  Pattern Recognition}, 2017.

\bibitem{shapematching}
Olga Sorkine-Hornung and Michael Rabinovich.
\newblock {Least-Squares Rigid Motion Using SVD}.
\newblock {\em Computing}, 2017.

\bibitem{spezialetti2020learning}
Riccardo Spezialetti, Federico Stella, Marlon Marcon, Luciano Silva, Samuele
  Salti, and Luigi Di~Stefano.
\newblock {Learning to Orient Surfaces by Self-supervised Spherical CNNs}.
\newblock {\em Advances in Neural Information Processing Systems}, 2020.

\bibitem{srivastava2019geometric}
Nitish Srivastava, Hanlin Goh, and Ruslan Salakhutdinov.
\newblock {Geometric Capsule Autoencoders for 3D Point Clouds}.
\newblock {\em arXiv Preprint}, 2019.

\bibitem{sun2020acne}
Weiwei Sun, Wei Jiang, Eduard Trulls, Andrea Tagliasacchi, and Kwang~Moo Yi.
\newblock {ACNe: Attentive Context Normalization for Robust
  Permutation-Equivariant Learning}.
\newblock In {\em Conference on Computer Vision and Pattern Recognition}, 2020.

\bibitem{keypointnet}
Supasorn Suwajanakorn, Noah Snavely, Jonathan~J Tompson, and Mohammad Norouzi.
\newblock {Discovery of Latent 3D Keypoints via End-to-End Geometric
  Reasoning}.
\newblock 2018.

\bibitem{thomas2018tensor}
Nathaniel Thomas, Tess Smidt, Steven Kearnes, Lusann Yang, Li Li, Kai Kohlhoff,
  and Patrick Riley.
\newblock {Tensor Field Networks: Rotation-and Translation-Equivariant Neural
  Networks for 3D Point Clouds}.
\newblock {\em arXiv Preprint}, 2018.

\bibitem{tulsiani2017learning}
Shubham Tulsiani, Hao Su, Leonidas~J Guibas, Alexei~A Efros, and Jitendra
  Malik.
\newblock Learning shape abstractions by assembling volumetric primitives.
\newblock In {\em Conference on Computer Vision and Pattern Recognition}, 2017.

\bibitem{conshier_sig13}
Oliver van Kaick, Kai Xu, Hao Zhang, Yanzhen Wang, Shuyang Sun, Ariel Shamir,
  and Daniel Cohen-Or.
\newblock Co-hierarchical analysis of shape structures.
\newblock {\em ACM SIGGRAPH}, 2013.

\bibitem{wang2019normalized}
He Wang, Srinath Sridhar, Jingwei Huang, Julien Valentin, Shuran Song, and
  Leonidas~J Guibas.
\newblock {Normalized Object Coordinate Space for Category-level 6d Object Pose
  and Size Estimation}.
\newblock In {\em Conference on Computer Vision and Pattern Recognition}, 2019.

\bibitem{wang2018pixel2mesh}
Nanyang Wang, Yinda Zhang, Zhuwen Li, Yanwei Fu, Wei Liu, and Yu-Gang Jiang.
\newblock Pixel2mesh: Generating 3d mesh models from single rgb images.
\newblock In {\em European Conference on Computer Vision}, 2018.

\bibitem{wang2017cnn}
Peng-Shuai Wang, Yang Liu, Yu-Xiao Guo, Chun-Yu Sun, and Xin Tong.
\newblock {O-CNN: Octree-Based Convolutional Neural Networks for 3d Shape
  Analysis}.
\newblock {\em ACM Transactions on Graphics}, 36(4):1--11, 2017.

\bibitem{wang2018adaptive}
Peng-Shuai Wang, Chun-Yu Sun, Yang Liu, and Xin Tong.
\newblock Adaptive o-cnn: A patch-based deep representation of 3d shapes.
\newblock {\em ACM Transactions on Graphics (TOG)}, 37(6):1--11, 2018.

\bibitem{wang2019deep}
Yue Wang and Justin~M Solomon.
\newblock {Deep Closest Point: Learning Representations for Point Cloud
  Registration}.
\newblock In {\em International Conference on Computer Vision}, 2019.

\bibitem{wang2019dynamic}
Yue Wang, Yongbin Sun, Ziwei Liu, Sanjay~E Sarma, Michael~M Bronstein, and
  Justin~M Solomon.
\newblock {Dynamic graph cnn for learning on point clouds}.
\newblock {\em ACM Transactions on Graphics}, 2019.

\bibitem{Wu18a}
Yuxin Wu and Kaiming He.
\newblock {Group Normalization}.
\newblock In {\em European Conference on Computer Vision}, 2018.

\bibitem{pointcontrast}
Saining Xie, Jiatao Gu, Demi Guo, Charles~R. Qi, Leonidas~J. Guibas, and Or
  Litany.
\newblock {PointContrast: Unsupervised Pre-training for 3D Point Cloud
  Understanding}.
\newblock In {\em European Conference on Computer Vision}, 2020.

\bibitem{yan2016perspective}
Xinchen Yan, Jimei Yang, Ersin Yumer, Yijie Guo, and Honglak Lee.
\newblock {Perspective Transformer Nets: Learning Single-View 3D Object
  Reconstruction without 3D Supervision}.
\newblock In {\em Advances in Neural Information Processing Systems}, 2016.

\bibitem{yang2018foldingnet}
Yaoqing Yang, Chen Feng, Yiru Shen, and Dong Tian.
\newblock {FoldingNet: Point Cloud Auto-Encoder via Deep Grid Deformation}.
\newblock In {\em Conference on Computer Vision and Pattern Recognition}, 2018.

\bibitem{yew2020rpm}
Zi~Jian Yew and Gim~Hee Lee.
\newblock {Rpm-net: Robust point matching using learned features}.
\newblock In {\em Conference on Computer Vision and Pattern Recognition}, 2020.

\bibitem{yuan2020deepgmr}
Wentao Yuan, Ben Eckart, Kihwan Kim, Varun Jampani, Dieter Fox, and Jan Kautz.
\newblock {DeepGMR: Learning Latent Gaussian Mixture Models for Registration}.
\newblock In {\em European Conference on Computer Vision}, 2020.

\bibitem{zhao20193d}
Yongheng Zhao, Tolga Birdal, Haowen Deng, and Federico Tombari.
\newblock {3D Point Capsule Networks}.
\newblock In {\em Conference on Computer Vision and Pattern Recognition}, 2019.

\bibitem{zhao2019quaternion}
Yongheng Zhao, Tolga Birdal, Jan~Eric Lenssen, Emanuele Menegatti, Leonidas
  Guibas, and Federico Tombari.
\newblock {Quaternion Equivariant Capsule Networks for 3D Point Clouds}.
\newblock In {\em European Conference on Computer Vision}, 2020.

\bibitem{zhou2016fast}
Qian-Yi Zhou, Jaesik Park, and Vladlen Koltun.
\newblock {Fast global registration}.
\newblock In {\em European Conference on Computer Vision}, 2016.

\end{thebibliography}
}

\clearpage
\appendix
{
\centering
\Large
\textbf{Canonical Capsules: Self-Supervised Capsules in Canonical Pose} \\
\vspace{0.5em}Supplementary Material \\
\vspace{1.0em}
}

This appendix provides additional architectural details~(\Section{arch_detail}), and additional ablation studies~(\Section{additional_abs}).
Note that we also provide code and additional qualitative results as a webpage with videos alongside this appendix; see~\texttt{README.html}.

\section{Architectural details}
\label{sec:arch_detail}

We detail our architecture design for the capsule encoder $\acne$, the decoder $\decoder$ and the regressor $\regressor$.

\subsection{Capsule Encoder -- $\acne$}
\label{sec:encoder_appendix}

The capsule encoder $\acne$ takes in input a point cloud $\points \in \real^{P\times D}$ and outputs the $K$-fold/head attention map $\attention \in \real^{P{\times}K}$ and the feature map  $\featuremap \in \real^{P\times C}$.
Specifically, the encoder $\acne$ is based on ACNe~\cite{sun2020acne} and composed of 3 residual blocks where each residual block consists of two hidden MLP layers with 128 neurons each.
Each hidden MLP layer is followed by \textit{Attentive Context Normalization} (ACN)~\cite{sun2020acne}, batch normalization~\citeA{Ioffe15}, and ReLU activation.
To be able to attend to multiple capsules, we extend the ACN layer into multi-headed ACN.

\paragraph{Multi-headed ACN}
For each MLP layer where we apply ACN, if we denote this layer as $i$, we first train a fully-connected layer that creates a $K$-headed attention map $\attention^i \in \real^{P{\times}K}$ given the $\featuremap^i \in \real^{P\times C}$ of this layer.
This is similar to ACN, but instead of a single attention map, we now have~$K$.
The normalization process is similar to ACN: we utilize the weighted moments of $\featuremap^i$ with $\attention^i$, but results in $K$ normalized outcomes instead of one.
We them aggregate these $K$ normalization results into one by summation.
In more details, given the $\featuremap_p^i \in \real^{P\times C}$, if we denote the weights and biases to be trained for the $k^{th}$ attention head to be $\bW_k^i \in \mathbb{R}^{C\times 1}$ and $b_k^i \in \mathbb{R}^{1}$ we write:
\begin{equation}
    \attention_{p, k}^i = \frac{\text{exp}(\featuremap_p^i \bW_k^i + b_k^i)}{\sum_k \text{exp}(\featuremap_p^i \bW_k^i + b_k^i )}
    \;.
\end{equation}
We then compute the moments that are used to normalize in ACN, but now for each attention head: 
\begin{equation}
\bmu_k = \sum_p \frac{\attention^i_{p, k} \featuremap^i_p}{\sum_p \attention^i_{p, k}}\;,\;
\bsigma_k = \sum_{p}\frac{\attention^i_{p, k} (\featuremap^i_p - \mu_k)^2}{\sum_p \attention^i_{p, k}} 
\;,
\end{equation}
which we then use to normalize and aggregate (sum) to get our final normalized feature map:
\begin{equation}
\featuremap^i_p = \sum_k  \attention^i_{p, k} \frac{(\featuremap^i_p - \bmu_k)}{\sqrt{\bsigma_k + \epsilon}}
\;,
\end{equation}
where $\epsilon=0.001$ is to avoid numerical instabilities.

\subsection{Capsule Decoder -- $\decoder$}
\label{sec:decoder_appendix}
The decoder $\decoder$ is composed of $K$ per-capsule decoders $\decoder_k$. 
Each per-capsule decoder $\decoder_k$ maps the $k^{th}$ capsule in canonical frame ($\bar\mR\pose_k+\bar\vt$, $\descriptor_k$) to a group of points $\tilde\points_k \in \real^{M\times D}$ that corresponds to a part of the entire object. 
We then obtain the auto-encoded (reconstructed) point clouds $\tilde\points$ by collecting the outputs of $K$ per-capsule decoders and taking their union as in \eq{recon}.
Specifically, each $\decoder_k$ consists of 3 hidden MLP layers of (1280, 640, 320) neurons, each followed by batch normalization and a ReLU activation.
At the output layer, we use a fully connected layer, followed by \texttt{tanh} activation to regress the coordinates for each point.
Similarly to \atlasnet, $\decoder_k$ additionally receives a set of trainable grids of size ($M\times 10$)\footnote{Note the constant $10$ is not related to $K$ here.} and deforms them into $\tilde\points_k$, based on the shape code~$\descriptor_k$.
Note that the generated point clouds from $\decoder_k$ lie in the reference frame of the canonicalized pose $\bar\pose_k$, so we rotate and translate the point cloud by $\bar\pose_k$ in the output layer.

\subsection{Regressor-- $\regressor$}
\label{sec:regressor_appendix}
{%
The regressor $\regressor$ learns to regress the canonical pose for each capsule $\bar\pose \in \real^{K\times D}$ from their descriptors $\descriptor_k$.
In order to do so, we concatenate the (ordered) set of descriptors $\{\descriptor_k\}$ into a single, global descriptor, which we then feed into a fully connected layer with $128 \times K$ neurons and a ReLU activation, followed by an additional fully connected layer with $D \times K$ neurons, creating $K$ $D$-dimensional outputs (i.e. canonical pose).
We do not apply any activation on the second layer.
To make $\bar\pose$ zero-centered in the canonical frame, we further subtract the mean of the outputs.
}

\def\capshift{-4}
\def \qualfigwidth {0.162}
\def \qualcapwidth {0.15}
\begin{figure*}
    \begin{center}
        \begin{overpic}
            [width=\linewidth]
            {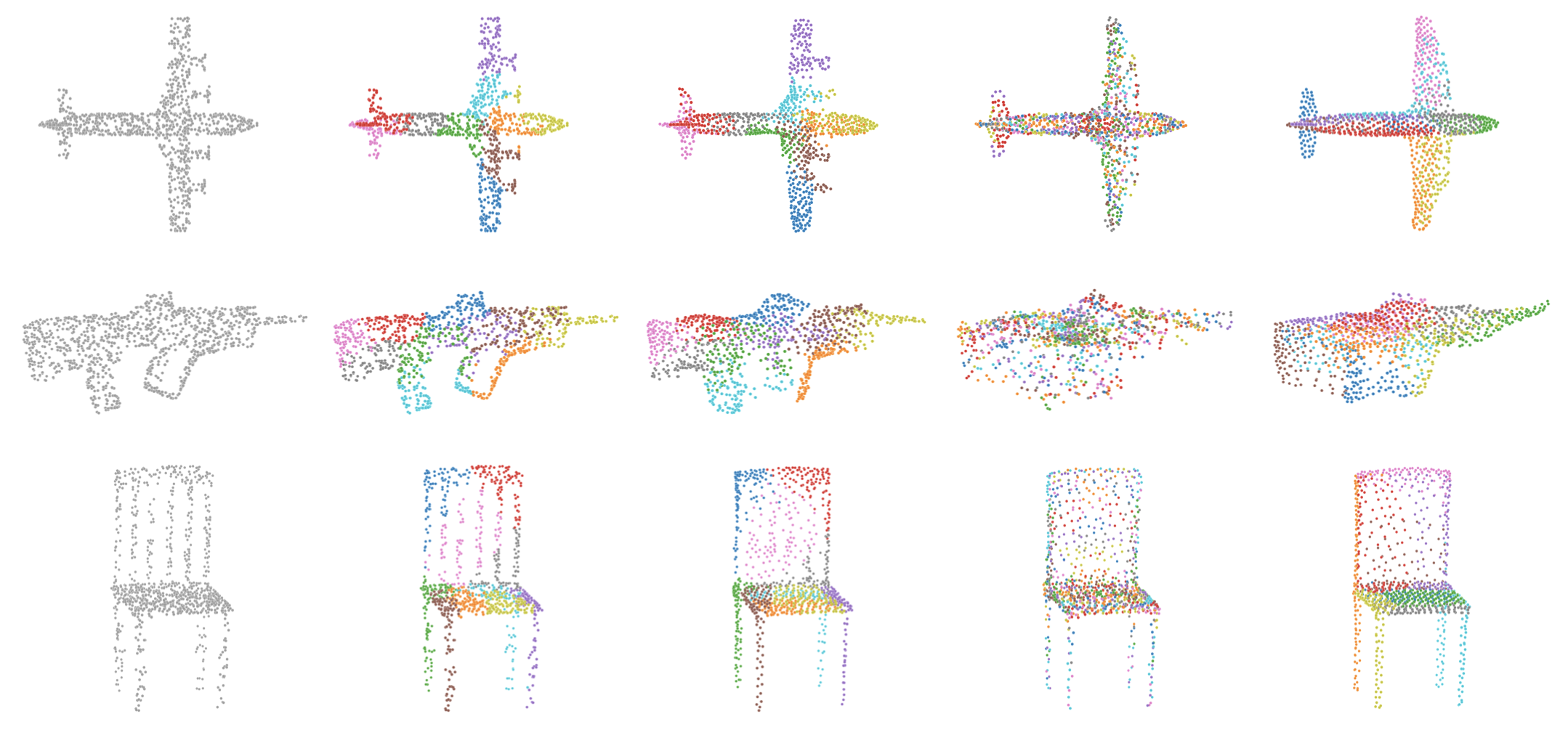}
            \put(3,\capshift){\parbox{\qualcapwidth\linewidth}{\centering \small Input}}
            \put(23,\capshift){\parbox{\qualcapwidth\linewidth}{\centering \small \textit{Our} capsule decomposition}}
            \put(43,\capshift){\parbox{\qualcapwidth\linewidth}{\centering \small \textit{Our} reconstruction}}
            \put(60,\capshift){\parbox{\qualcapwidth\linewidth}{\centering \small \capsnet reconstruction}}
            \put(83,\capshift){\parbox{\qualcapwidth\linewidth}{\centering \small \atlasnet reconstruction}}
        \end{overpic}
        \vspace{1em}
    \end{center}
    \caption{
        \textbf{Auto-encoding / qualitative -- }
        Example decomposition results using Canonical Capsules on the test set, with the \textit{aligned} setup; we color each decomposition (capsule) with a unique color. For \capsnet and \atlasnet, rather than capsules, these correspond to ``patches'' in the reconstruction network. Our method clearly provides the best qualitative results.
    }
    \label{fig:qualitative_canonical}
\end{figure*}

\section{Additional ablation studies}
\label{sec:additional_abs}

\paragraph{Auto-encoding with aligned data -- \Figure{qualitative_canonical} and \Table{both_recons}}
For completeness, we further show qualitative results for auto-encoding on an \textit{aligned} dataset, the most common setup of prior works in the literature.
As shown in \Figure{qualitative_canonical}, our method provides best reconstruction performance even in this case; for quantitative results, please see~\Table{both_recons}. 
Interestingly, while our \textit{decoder} architecture is similar to \atlasnet, our reconstructions are of much higher quality; our methods provides finer details at the propellers on the airplane, at the handle of the firearm, and at the back of the chair.
This further supports the effectiveness of our capsule encoding.

\begin{wrapfigure}[16]{R}{0.40\linewidth}
    \vspace{-1em}
    \captionof{table}{
        \textbf{Number of points $P$ -- }
        Auto-encoding performance~(Chamfer distance $\times$ $10^3$) as we vary the input point cloud cardinality; $\it aligned$ setup for both training and testing, with \textit{all} object categories.
    }
    \begin{center}
    \small
    \resizebox{\linewidth}{!}{
        \begin{tabular}{lcc}
            \toprule
                                                   & 1024 pts      & 2500 pts      \\
            \midrule
            \capsnet                               & 2.49          & 1.49          \\
            AtlasNetV2~\cite{deprelle2019learning} & 2.14          & 1.22          \\
            Our method                             & \textbf{1.76} & \textbf{0.97} \\
            \bottomrule
        \end{tabular}
    } %
    \end{center}
    \label{tbl:ablation_numpt}
\end{wrapfigure}

\paragraph{Number of points $P$ -- \Table{ablation_numpt}}
To speed-up experiments we have mostly used $P{=}1024$, but in the table we show that our findings are consistent regardless of the number of points used.
Note that the results of \atlasnet are \textit{very} similar to what is reported in the original paper.
The slight differences exist due to random subsets that were used in AtlasNetV2.\footnote{And to a minor bug in the evaluation code (\ie~non deterministic test set creation) that we have already communicated to the authors of~\cite{groueix2018papier}.}

\begin{table}
    \begin{minipage}{0.58\linewidth}
\captionof{table}{
\textbf{Ablation study on the number of capsules --} 
We show the reconstruction performance (Chamfer distance $\times$ $10^3$) with varying number of capsules.
While they all perform better than competitors, 10 capsules give best performance.
Note the representation power is kept \textit{constant} as we vary the number of capsules.
}
\begin{center}
\resizebox{\linewidth}{!}{
\begin{tabular}{cccc}
\toprule
          \atlasnet   & 5 capsules & 10 capsules & 20 capsules \\
\midrule
  2.80 & 1.25           & \textbf{1.11}        & 1.15        \\
\bottomrule
\end{tabular}}
\end{center}
\label{tbl:nums_capsules}
\end{minipage}
\hfill
    \begin{minipage}{0.40\linewidth}
\captionof{table}{
\textbf{Number of capsules w/ fixed descriptor dimension per capsule -- } 
Better reconstruction performance is obtained as number of capsules are increased, since the network capacity expands proportionally.
}
\begin{center}
\resizebox{\linewidth}{!}{
\begin{tabular}{lccc}
\toprule
Number of capsules & 5    & 10   & 20   \\
\midrule
CD                 & 1.51 & 1.11 & 1.02 \\
\bottomrule
\end{tabular}}
\end{center}
\label{tbl:fixed_dim}
\end{minipage}

\end{table}

\paragraph{Effect of number of capsules -- \Table{nums_capsules}}
To verify how the number of capsules affect performance, we test with varying the number of capsules.
We keep the representation power constant by \textit{reducing} the dimension of descriptors as more capsules are used; for example, with 10 capsules we use a 128-dimensional descriptor, with 20 we use 64, and with 5, we use 256.
Our experimental results show that representation with 10 capsules achieves the best performance.
Note that our method, even with the sub-optimal number of capsules, still outperforms compared methods by a large margin.  

\paragraph{Increasing the number of capsules w/ fixed descriptor dimension -- \Table{fixed_dim}}
We also report the performance for the increasing number of capsules while keeping descriptor dimension fixed. 
Note this is different from \Table{nums_capsules}, as in that setting the overall network capacity was kept fixed, but here it grows linearly to the number of capsules.
As shown in the \Table{fixed_dim}, unsurprisingly, more capsules lead to better reconstruction performance as the network capacity is enhanced.

\begin{table}
    \begin{minipage}{0.45\linewidth}
\captionof{table}{
\textbf{One-shot canonicalization --} 
Reconstruction performance of our method against a naive one-shot alignment, where an arbitrary point cloud is selected as reference.
}
\begin{center}
\resizebox{\linewidth}{!}{
\begin{tabular}{lccc}
\toprule
                      & Airplane & Chair & All \\
\midrule
One shot alignment  & \textbf{1.10} & 2.92 &2.37         \\         
Our method & 1.11 & \textbf{2.58} &\textbf{2.22}          \\
\bottomrule
\end{tabular}
}
\end{center}
\label{tbl:ablation_regressor}
\end{minipage}
\hfill
    \begin{minipage}{0.50\linewidth}
\captionof{table}{
\textbf{Canonical descriptors -- }
Auto-encoding performance~(Chamfer Distance) of our method with descriptors 
$\descriptor_k$ vs. $\bar\descriptor_k$.
Note the unaligned setup is the one of primary interest.
}
    \begin{center}
        \resizebox{\linewidth}{!}{
            \begin{tabular}{lccc}
            \toprule
            & AtlasNetV2~\cite{deprelle2019learning} & Ours ($\descriptor_k$) & Ours ($\bar\descriptor_k$)\\
            \midrule
            Aligned & 1.28 & \textbf{0.96} & {0.99} \\
            Unaligned & 2.80 & 2.12 & \textbf{1.08}          \\
            \bottomrule
            \end{tabular}
        } %
    \end{center}
    \label{tbl:ablation_desc}
\end{minipage}
 
\end{table}

\paragraph{One-shot canonicalization -- \Table{ablation_regressor}}
\label{sec:ablation_regressor}
A naive alternative to our learnt canonicalizer would be to use one point cloud as a reference to align to.
Using the canonicalizer provides improved reconstruction performance over this na\"ive approach, removes the dependency on the choice of the reference point cloud, and allows our method to work effectively when dealing with multi-class canonicalization.

\paragraph{Canonical descriptors -- \Table{ablation_desc}}
\label{sec:ablation_desc}
We evaluate the effectiveness of the descriptor enhancement strategy described in~\Section{arch}.
We report the reconstruction performance with and without the enhancement.
Recomputing the descriptor in canonical frame helps when dealing with the $\it unaligned$ setup.
Note that even without this enhancement, our method still outperforms the state-of-the-art.

\begin{wrapfigure}{r}{0.5\linewidth}
\vspace{-1em}
\captionof{table}{
\textbf{The impact of $\loss{canonical}$ on canonicalization --} 
Best canonicalization is achieved when $\loss{canonical}$ is used; see text for details.
}
\begin{center}
\resizebox{\linewidth}{!}{
\begin{tabular}{lcccc}
\toprule
    & \multicolumn{2}{c}{w/o $\loss{recon}$} &\multicolumn{2}{c}{w/ $\loss{recon}$} \\ 
                                                        \cmidrule(r){2-3}
                                                        \cmidrule(r){4-5}
   &w/o $\loss{canonical}$ & w/ $\loss{canonical}$ & w/o $\loss{canonical}$ & w/ $\loss{canonical}$ \\
\midrule
CD    & -   & - & 1.12   & \textbf{1.11}  \\ 
mStd    &16.0  &\textbf{7.747} &8.421    & \textbf{8.278}  \\
\bottomrule
\end{tabular}}
\end{center}
\label{tbl:canonical_loss}
\end{wrapfigure}
\paragraph{The impact of $\loss{canonical}$}
Using $\loss{canonical}$ is required to achieve the best performance; see the inset table.
However, the reconstruction loss $\loss{recon}$ alone is able to guide canonicalization up to some degree.
It is worth noting that excluding $\loss{recon}$, thus training the canonicalization component in a standalone fashion, unsurprisingly provides best mStd performance.
Thus, this could be an alternative strategy to train our framework, should one \textit{not} require end-to-end differentiability. 
Nonetheless, our observations still hold and our method delivers the best performance (including when compared to \compass) with all losses enabled.

\newcommand{\imrrwidth}{.32\linewidth}
\begin{figure}[t]
    \begin{center}
        \subfigure[]{\includegraphics[width=\imrrwidth]{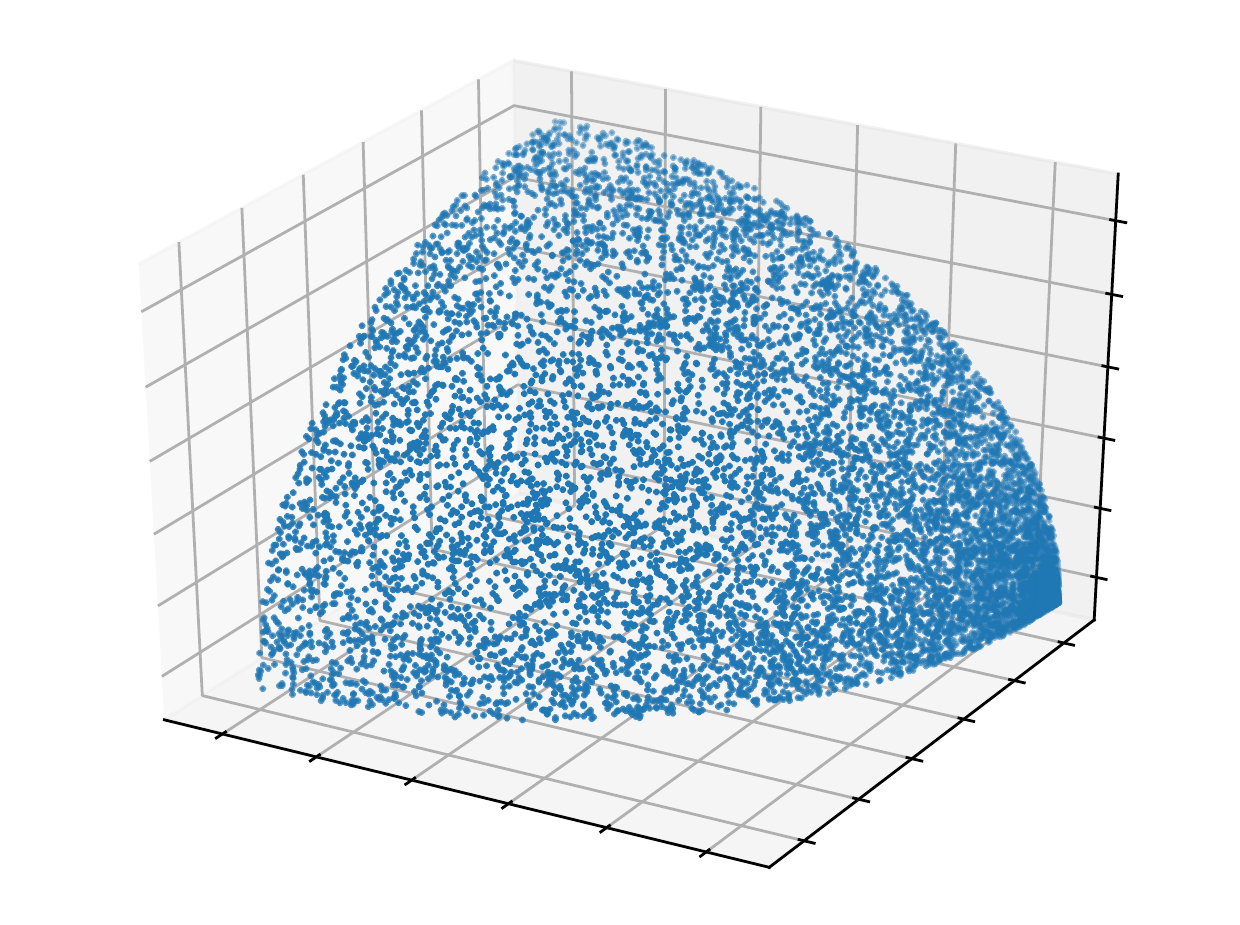}}
        \subfigure[]{\includegraphics[width=\imrrwidth]{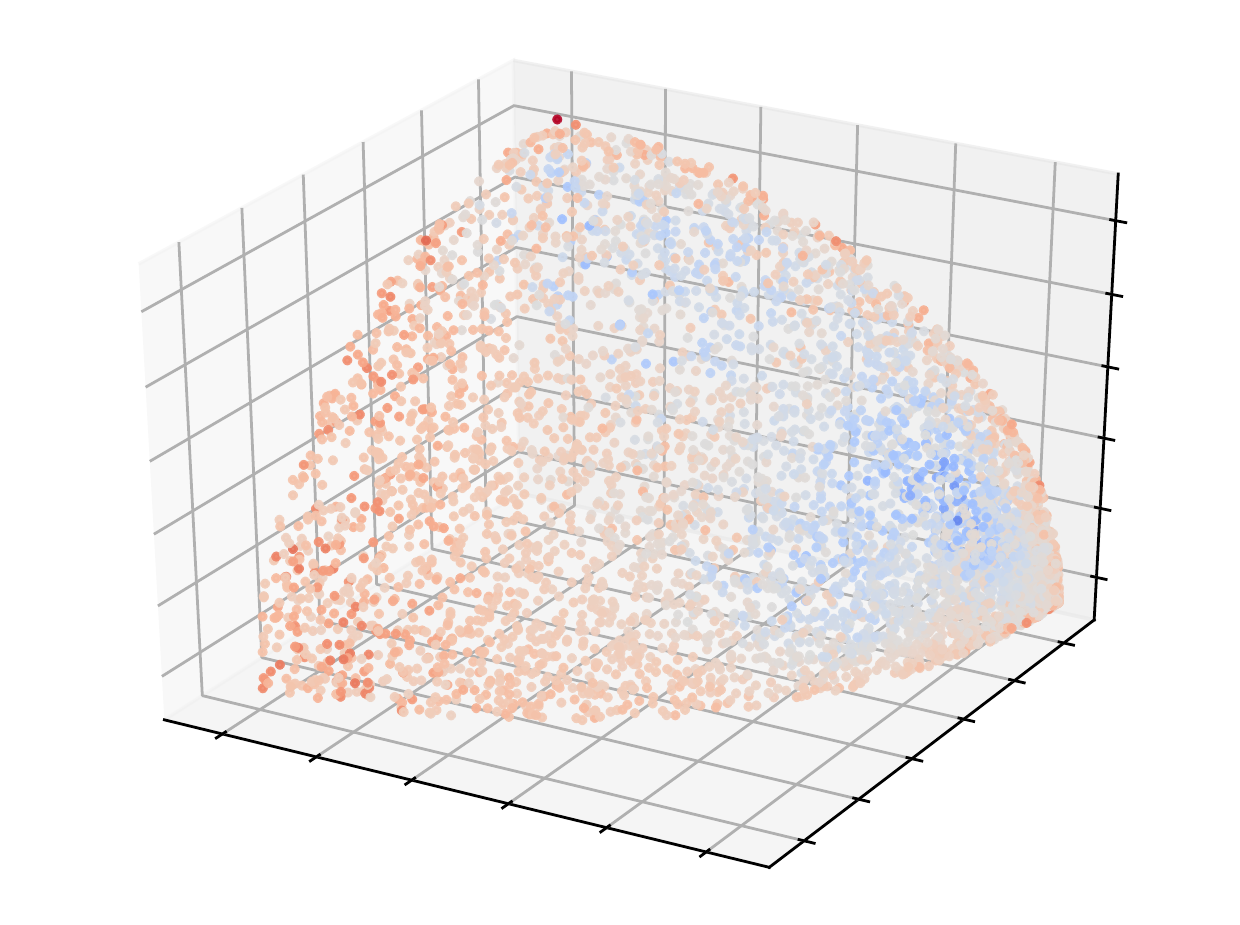}}
        \subfigure[]{\includegraphics[width=\imrrwidth]{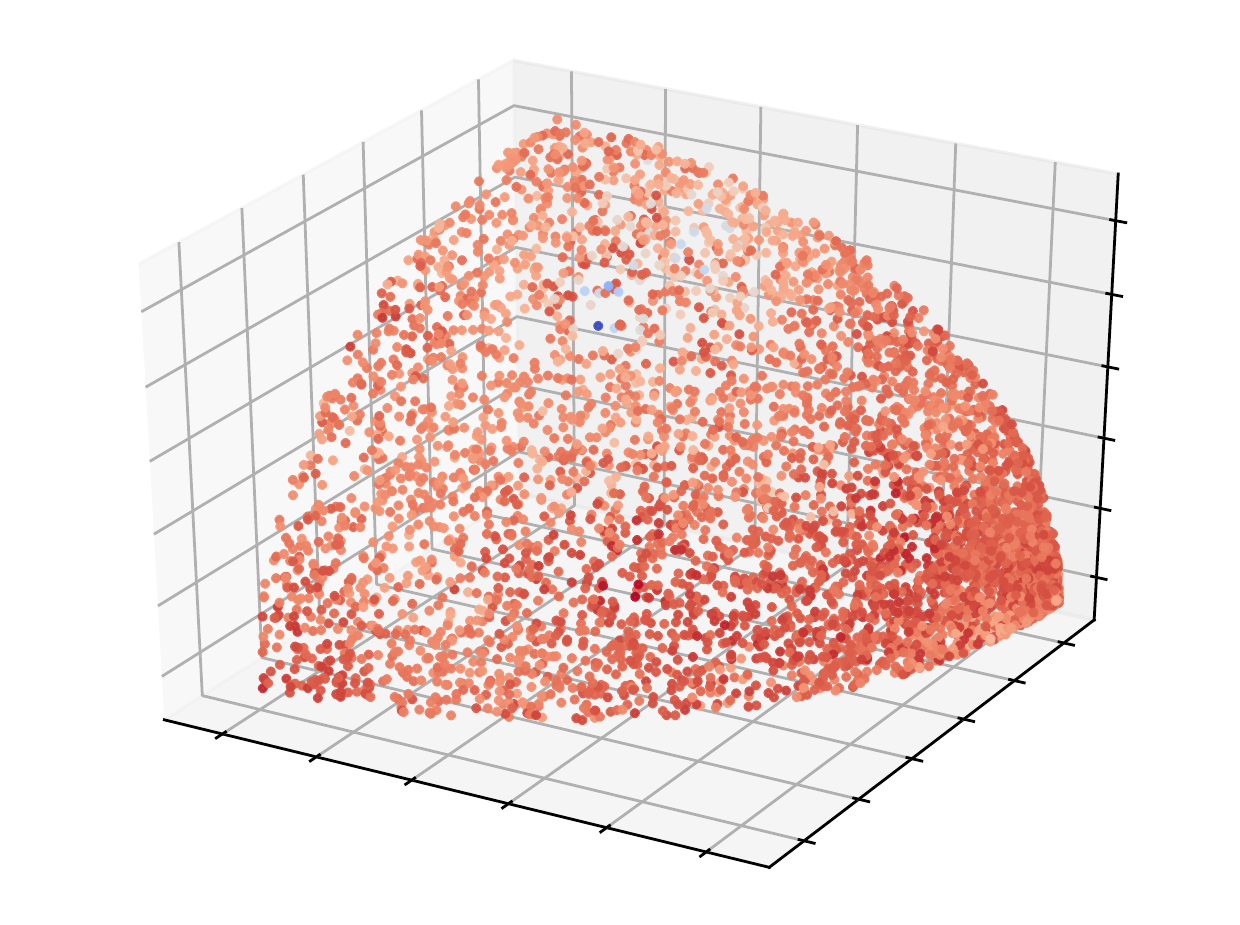}}
    \end{center}
    \vspace{-0.5em}
    \caption{
        \textbf{Random sampling of rotations -- }
        (a) Sampling Euler angles uniformly results in a non-uniform coverage of SO(3); note we visualize only one-eighth of a sphere for ease of visualization on paper.
        (b) Non uniform sampling results in auto-encoding error to be biased w.r.t rotations (we use a cold-warm colormap to visualize the Chamfer distance error).
        (c) By properly sampling rotations~\cite{shoemake1992uniform}, this bias can be alleviated.
    }
    \label{fig:random_rot}
\end{figure}

\paragraph{Supervising the attention's invariance}
Since $\pose$ is inferred by weighted averaging of $\points$ with the attention map $\bA$ in \eq{capsule}, we also considered directly adding a loss on $\bA$ that enforces invariance instead of a loss on $\pose$.
This variant degrades only slightly in terms reconstruction ($CD{=}1.13$, where ours provides $CD{=}\bf 1.11$) but performs very poorly when canonicalizing ($\text{mStd}{=}94.906$ vs $\text{mStd}{=}\bf 8.278$ of ours).
We hypothesize that this is because $\loss{equivariance}$ \textit{directly} supervises the end-goal~(capsule pose equivariance) whereas supervising $\bA$ is an indirect one.

\paragraph{Random sampling of rotations -- \Figure{random_rot}}
\label{sec:ablation_rot}
Lastly, we revisit how rotations are randomly sampled to generate the augmentations used by Siamese training.
Uniform sampling of Euler angles~(i.e., yaw,pitch,roll) leads to a non-uniform coverage of the SO(3) manifold as shown in~\Figure{random_rot}~(a).
Due to this non-uniformity, the reconstruction quality is biased with respect to test-time rotations; see~\Figure{random_rot}~(b).
Instead, by properly sampling~\cite{shoemake1992uniform} the reconstruction performance is much more \textit{uniform} across the manifold; see see~\Figure{random_rot}~(c)
In our experiments, this leads to a significant difference in auto-encoding performance; $CD{=}\bf 1.11$ with proper uniform sampling vs $CD{=}1.19$ with the Euclidean random sampling.

\begin{wrapfigure}[6]{r}{0.45\linewidth}
\vspace{-1.2em}
\captionof{table}{
\textbf{Large range of random translations --} 
The performance changes are negligible as the magnitude of random translations increases.
}
\begin{center}
\vspace{-1.em}
\resizebox{\linewidth}{!}{
\begin{tabular}{lcccc}
\toprule
Range of translation & $[-0.2, 0.2]$ & $[-0.4, 0.4]$ & $[-0.8, 0.8]$ & $[-1.6, 1.6]$ \\
\midrule
CD                   & 1.11            & 1.12            & 1.1             & 1.1     \\
\bottomrule
\end{tabular}}
\end{center}
\label{tbl:large_translation}
\end{wrapfigure}

\paragraph{Large range of random translations -- \Table{large_translation}}
We increase the range of random translations ($[-0.2, 0.2]$ in the original paper). 
As shown in the \Table{large_translation}, we observe the negligible changes in reconstruction performance (CD) with larger random translations.

\section{Per-class results for auto-encoding}
In addition to the auto-encoding results in \Table{both_recons}, we provide per-class performance for the models trained with multiple categories.
As shown in \Table{perclass_recon}, we achieve the best performance for all classes.    
\begin{wrapfigure}[11]{r}{1.0\linewidth}
    \captionof{table}{
        \textbf{Auto-encoding / per-class quantitative --} 
        Performance in terms of Chamfer distance with 1024 points per point cloud -- metric is multiplied by $10^3$ as in \cite{deprelle2019learning}.
    } %
    \vspace{-.7em}
    \begin{center}
        \resizebox{\linewidth}{!}{
            \begin{tabular}{lccccccccccccccc}
                \toprule
                 & Method & Bench         & Cabinet       & Car           & Cellphone     & Chair         & Couch         & Firearm       & Lamp          & Monitor       & Airplane      & Speaker       & Table         & Watercraft    & All           \\
                \midrule
                \multirow{3}{*}{\rotatebox[origin=c]{90}{\footnotesize Aligned}}
                & \capsnet   & 2.06          & 3.23          & 2.64          & 2.25          & 2.64          & 2.99          & 0.93          & 3.40          & 2.85          & 1.36          & 4.26          & 2.56          & 2.05          & 2.49          \\
                & \atlasnet & 1.67          & 2.81          & 2.42          & 2.00          & 2.26          & 2.63          & 0.78          & 2.68          & 2.52          & 1.18          & 3.80          & 2.16          & 1.73          & 2.14          \\
                & Our method        & \textbf{1.44} & \textbf{2.37} & \textbf{2.10} & \textbf{1.77} & \textbf{1.90} & \textbf{2.26} & \textbf{0.59} & \textbf{1.61} & \textbf{2.09} & \textbf{0.99} & \textbf{3.04} & \textbf{1.80} & \textbf{1.31} & \textbf{1.76} \\
                \midrule
                \multirow{4}{*}{\rotatebox[origin=c]{90}{\footnotesize Unaligned}}
                & \capsnet   & 4.34          & 5.20          & 4.57          & 3.87          & 5.55          & 5.33          & 2.00          & 4.97          & 4.30          & 3.24          & 6.05          & 5.17          & 3.38          & 4.66          \\
                & \atlasnet & 2.93          & 3.65          & 3.46          & 2.46          & 3.53          & 3.72          & 1.28          & 2.91          & 3.06          & 2.17          & 4.42          & 3.20          & 2.27          & 3.08          \\
                & \atlasnet w/ STN & 2.44          & 3.30          & 3.10          & 2.17          & 2.93          & 3.28          & 0.96          & 2.56          & 2.70          & 1.60          & 3.96          & 2.67          & 1.86          & 2.60          \\
                & Our method & \textbf{1.84} & \textbf{2.72} & \textbf{2.45} & \textbf{1.86} & \textbf{2.72} & \textbf{2.70} & \textbf{0.67} & \textbf{2.20} & \textbf{2.46} & \textbf{1.30} & \textbf{3.55} & \textbf{2.36} & \textbf{1.53} & \textbf{2.22} \\
                \bottomrule
            \end{tabular}
        } %
    \end{center}
    \label{tbl:perclass_recon}
\end{wrapfigure}

\end{document}